\documentclass{article}
\PassOptionsToPackage{table}{xcolor}
\usepackage{iclr2027_conference,times}
\usepackage{graphicx}
\usepackage{adjustbox}
\usepackage{placeins}

\usepackage{booktabs}
\usepackage{amssymb}
\usepackage{multirow}
\usepackage{subfig}
\usepackage[table]{xcolor}
\definecolor{color-tab}{rgb}{0.95,0.95,0.95}

\usepackage{amsmath,amsfonts,bm}

\def\eqref#1{equation~\ref{#1}}
\def\1{\bm{1}}

\DeclareMathAlphabet{\mathsfit}{\encodingdefault}{\sfdefault}{m}{sl}
\SetMathAlphabet{\mathsfit}{bold}{\encodingdefault}{\sfdefault}{bx}{n}

\usepackage{hyperref}
\usepackage{url}

\title{Hihaze: One-Step Diffusion for\\
High-Resolution Real-World Image Dehazing}

\author{Zheng Chen$^{1}$\thanks{Equal contribution.},\enspace
Yuchen Huang$^{1}$\footnotemark[1],\enspace
Xiaoyang Liu$^{1}$,\enspace 
\textbf{Jue Gong}$^{1}$,\enspace
\textbf{Linghe Kong}$^{1}$,\enspace
\textbf{Yulun Zhang}$^{1}$\thanks{Corresponding author: Yulun Zhang, yulun100@gmail.com.} \\
\textsuperscript{1}Shanghai Jiao Tong University
\vspace{-5.mm}
}

\iclrfinalcopy
\begin{document}

\maketitle
\suppressfloats[t]

\begin{abstract}
Diffusion models have recently achieved encouraging results in real-world image dehazing. However, restoring high-resolution images remains challenging due to spatially non-uniform haze and inconsistent predictions across local patches. To address these issues, we propose Hihaze, a one-step diffusion framework for high-resolution real-world dehazing. Hihaze adopts two haze-aware modulation modules: attentive skip compensation (ASC) and a spatial degradation adapter (SDA). ASC selectively transfers reliable encoder information to the variational autoencoder (VAE) decoder, preserving structural details while suppressing haze-degraded features. SDA extracts multi-scale spatial degradation representations and injects them into the U-Net to guide restoration according to local haze variations. Furthermore, we design a training-free global-guided patch fusion (GPF) method. It uses a global prediction to align patch appearance and downweight unreliable predictions, followed by structure-guided detail compensation. Notably, this training-free design also makes GPF a plug-and-play module for other dehazing models. Experiments across four real-world benchmarks show that Hihaze surpasses state-of-the-art approaches in both visual quality and inference speed. Specifically, it delivers PSNR improvements of \textbf{3.62~dB} on Dense-Haze and \textbf{3.69~dB} on NH-HAZE compared with the strongest baselines. Code: \url{https://github.com/zhengchen1999/Hihaze}.
\end{abstract}
\setlength{\abovedisplayskip}{2pt}
\setlength{\belowdisplayskip}{2pt}
\section{Introduction}
\label{sec:intro}
Single-image dehazing reconstructs haze-free scenes from hazy inputs. Haze reduces contrast, shifts colors, and obscures details with spatially varying density. Physically, this degradation is formulated via the atmospheric scattering model~\citep{narasimhan2002vision,mccartney1977optics} 
\begin{equation}
    \mathbf{I}(x) = \mathbf{I}_{\mathrm c}(x)\,t_{\mathrm h}(x)
    + \mathbf{A}\bigl(1 - t_{\mathrm h}(x)\bigr),
    \label{eq:asm}
\end{equation}
where $\mathbf{I}$ and $\mathbf{I}_{\mathrm c}$ denote the hazy image and its clean counterpart. $\mathbf{A}$ is the atmospheric light, and $t_{\mathrm h}(x)$ is the transmission at pixel $x$. Recovering $\mathbf{I}_{\mathrm c}$ is ill-posed because the scattering parameters are unknown. Early methods use statistical priors, such as dark-channel sparsity~\citep{he2010single}, to constrain restoration. However, these assumptions may fail in complex scenes. Later CNN- and Transformer-based methods learn restoration mappings from data~\citep{cai2016dehazenet,liu2019griddehazenet,qin2020ffa,guo2022image,song2022vision}. Despite this progress, recovering realistic textures remains difficult under dense, spatially varying haze.

Recently, large pretrained diffusion models have provided rich generative priors for image restoration~\citep{wang2024exploiting,liu2024diff}. They help recover realistic textures from degraded inputs. For dehazing, existing approaches exploit these priors through realistic haze synthesis, unpaired training, and physical or textual guidance~\citep{wang2025learning,lan2025exploiting}. To reduce computation, latent diffusion performs denoising in a compact latent space~\citep{rombach2022high}. Further reducing the number of denoising steps accelerates inference and makes diffusion-based dehazing more practical~\citep{song2020denoising,wang2025learning}.

Nevertheless, two difficulties remain in high-resolution real-world dehazing:
\textbf{(1) Spatially non-uniform haze.} VAE compression can discard fine details. Encoder--decoder skip connections help retain input information~\citep{lan2025exploiting}. However, encoder features contain both scene structures and haze degradation. Directly transferring them may leave residual haze in the result. Moreover, high-level semantic guidance does not explicitly describe local degradation. It therefore provides limited guidance for spatially varying restoration. Effective dehazing requires reliable detail transmission together with spatial degradation guidance.
\textbf{(2) Inconsistent patch predictions.} High-resolution inference often uses patches to limit memory use~\citep{wang2024exploiting}. Each patch sees only part of the scene. Adjacent patches may therefore differ in color, illumination, and dehazing strength. Overlapping inference and local aggregation can reduce boundary artifacts~\citep{ozdenizci2023restoring,ding2024patched}. However, they lack an explicit global reference. Averaging inconsistent predictions may also blur fine textures.

\begin{figure}[t]
    \centering
    \includegraphics[width=1.\linewidth]{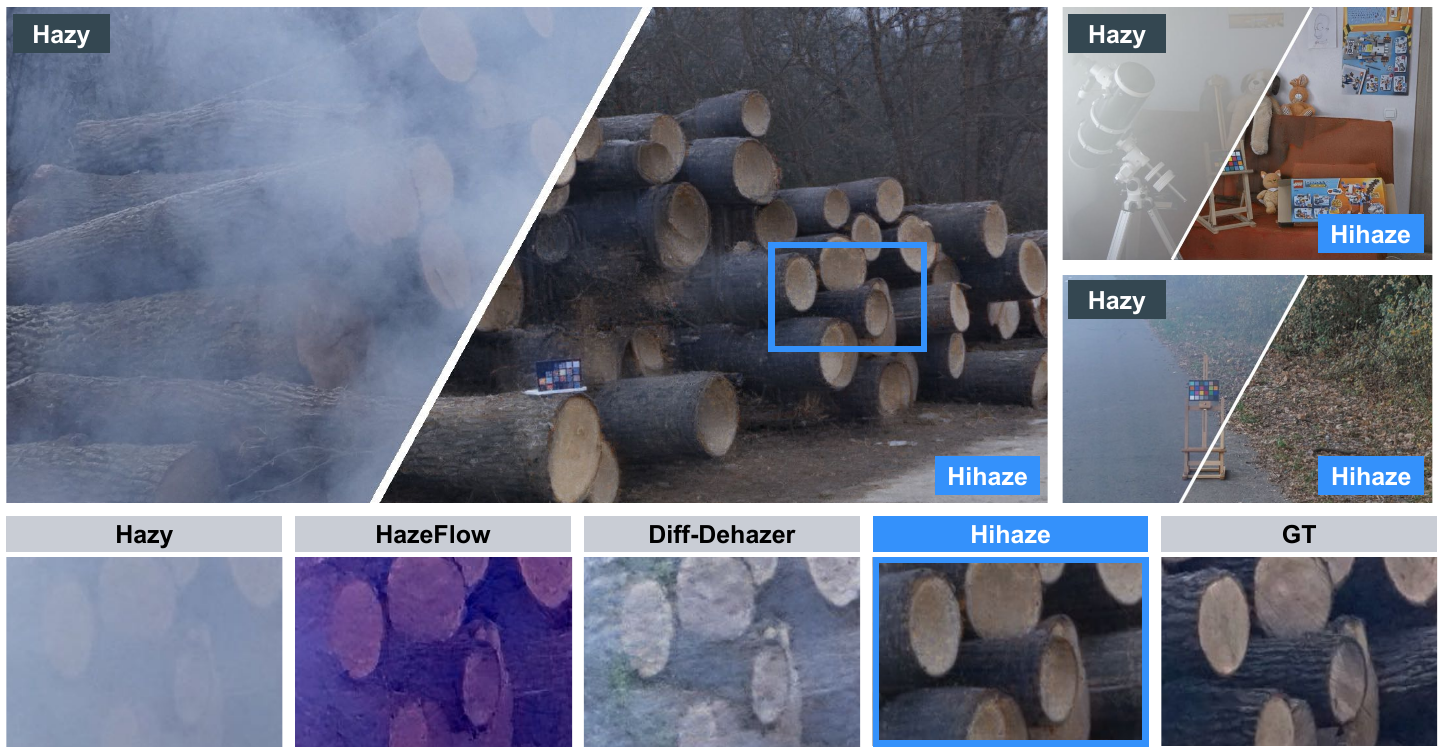}
    \caption{Visual results on NH-HAZE 51 (left), I-HAZE 34 (upper right),
    and O-HAZE 01 (lower right). Each scene shows the hazy input on the
    left of the diagonal and the Hihaze result on the right.
    Bottom: matched details from the marked region of NH-HAZE 51.
    Hihaze removes haze while recovering scene structures and textures in the restored images.}
    \label{fig:teaser}
    \vspace{-2mm}
\end{figure}

To address these issues, we propose \textbf{Hihaze} for high-resolution image dehazing. Hihaze adapts a pretrained latent diffusion model to restore hazy images in one step. We introduce two haze-aware modulation modules: attentive skip compensation (ASC) and a spatial degradation adapter (SDA). \textbf{(1) ASC} uses decoder features to retrieve reliable encoder information. It compensates for missing structures and fine details while suppressing haze-degraded features. \textbf{(2) SDA} extracts spatial degradation features from the hazy input. It injects these features into the U-Net at multiple scales. This guides restoration according to local haze variations.

Moreover, we design training-free global-guided patch fusion (GPF) for high-resolution inference. A prediction from the downsampled full image provides a shared appearance reference. GPF aligns local patch predictions with this reference and reduces the contribution of inconsistent patches. Structural agreement between the fused result and the hazy input then guides detail compensation. This improves cross-patch consistency while preserving reliable textures. GPF can also be integrated into other dehazing models as a plug-and-play module.

Extensive experiments on four real-world benchmarks demonstrate the effectiveness of Hihaze. As shown in Fig.~\ref{fig:teaser}, Hihaze removes more haze than the compared methods. It also recovers clearer edges and textures, even in severely degraded regions. Our main contributions are as follows:
\begin{itemize}
\item We propose Hihaze, a one-step diffusion dehazing framework with attentive skip compensation and a spatial degradation adapter for high-resolution images.
\item We introduce global-guided patch fusion to improve consistency and preserve fine textures. It is a training-free, plug-and-play module for other dehazing models.
\item We validate Hihaze on four real-world benchmarks. It achieves superior reconstruction fidelity and perceptual quality under diverse haze conditions.
\end{itemize}

\section{Related Work}
\subsection{Image Dehazing}
Early dehazing methods typically derive physical priors from natural-image statistics. Such constraints are subsequently coupled with the physical scattering formulation to restore haze-free images~\citep{fattal2008single,he2010single,meng2013efficient,fattal2014dehazing,berman2016non}. Specifically, \citet{he2010single} developed the Dark Channel Prior (DCP), noting that the minimum intensity across RGB channels within a haze-free local patch typically approaches zero. Similarly, \citet{fattal2014dehazing} proposed the Color-Line Prior, leveraging the one-dimensional distributions formed by pixels in local image patches in RGB color space to estimate scene transmission. However, handcrafted priors often struggle to model complex haze in real-world scenes, potentially causing color distortions.

Driven by advances in deep learning, single-image dehazing increasingly relies on deep neural networks. With task-specific architectures, these models directly map hazy inputs to haze-free outputs~\citep{liu2019griddehazenet,dong2020multi,guo2022image,song2022vision}. Although effective on synthetic benchmarks, these methods often generalize poorly to real-world scenes, where haze distributions are considerably more complex and diverse. To improve real-world generalization, recent works incorporate external priors or adaptive refinement mechanisms into the restoration pipeline~\citep{wu2023ridcp,fang2024real,chen2024prompt,fu2025iterative}. For example, RIDCP~\citep{wu2023ridcp} leverages high-quality codebook priors to guide the restoration of real-world hazy images. More recently, IPC-Dehaze~\citep{fu2025iterative} employs an iterative predictor-critic framework that retains reliable codes and progressively refines code predictions. Nevertheless, constrained by limited generative capacity, standard CNN and Transformer architectures struggle to recover realistic details under heavy, spatially varying haze.

\subsection{Diffusion Model-Based Image Restoration}
Recent breakthroughs in diffusion models have shown remarkable capacity for complex image generation~\citep{song2020denoising,ho2020denoising,rombach2022high,ruiz2023dreambooth}. By capturing rich priors of natural scenes, these generative frameworks have been widely adopted across various restoration tasks with compelling results~\citep{saharia2022image,saharia2022palette,lugmayr2022repaint,zheng2024selective,liu2024diff,chen2026diffst,chen2026dvface}. In the context of real-world image dehazing, DiffDehaze~\citep{wang2025learning} combines realistic haze synthesis with accelerated fidelity-preserving sampling to achieve efficient and faithful restoration. Diff-Dehazer~\citep{lan2025exploiting} combines unpaired bidirectional diffusion mapping with physical priors to improve real-world generalization. Despite these advances, existing methods remain limited in handling spatially non-uniform haze in real-world scenes, often resulting in uneven restoration across image regions and residual haze.

Moreover, the high memory cost of diffusion models often necessitates patch-wise inference for high-resolution images, resulting in inconsistent predictions across patches and visible boundary artifacts~\citep{wang2025learning,lan2025exploiting}. Existing approaches mitigate boundary artifacts through overlapping patch inference and local aggregation~\citep{ozdenizci2023restoring,ding2024patched,wang2024exploiting}. Specifically, Patch-DM~\citep{ding2024patched} suppresses seam artifacts by assembling features across adjacent regions for shifted-patch prediction.
However, such aggregation mainly enforces consistency between neighboring patches. Averaging predictions in overlapping regions may also attenuate fine high-frequency textures in the restored images. In this paper, we introduce a one-step diffusion framework for high-resolution real-world image dehazing that enables memory-efficient patch-wise inference and effectively handles non-uniform haze.

\section{Method}
\label{sec:method}
\begin{figure*}[t]
    \centering
    \vspace{-3mm}
    \includegraphics[width=\textwidth]{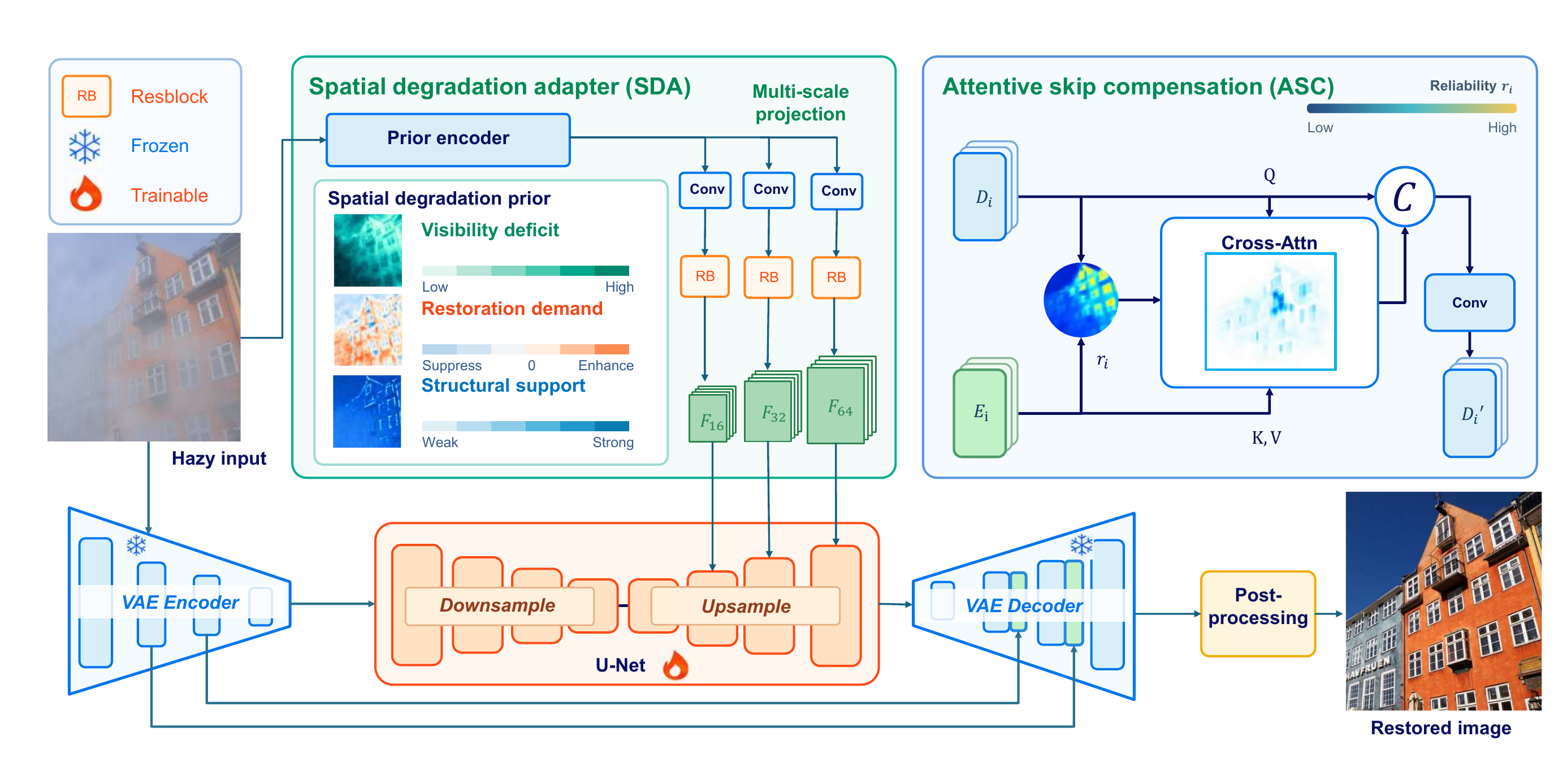}
    \caption{Overview of Hihaze. SDA guides spatially adaptive denoising,
    while ASC transfers reliable encoder features to the VAE decoder.
    GPF combines global and local predictions for high-resolution restoration with consistent appearance and preserved fine details.}
    \label{fig:example}
    \vspace{-2mm}
\end{figure*}
This section details Hihaze, our one-step diffusion approach designed to restore high-resolution hazy scenes.
We first introduce the overall framework in Sec.~\ref{sec:method-1}.
We then describe attentive skip compensation (ASC) in Sec.~\ref{sec:asc}
and the spatial degradation adapter (SDA) in Sec.~\ref{sec:sda}.
ASC retrieves reliable encoder features for VAE decoding, while SDA guides
denoising according to local haze variations.
Finally, Sec.~\ref{sec:method-3} presents training-free global-guided patch fusion
(GPF) for appearance consistency and structure-guided detail compensation.
Training details are provided in Appendix~\ref{app:training_details}.

\subsection{Overall Framework}
\label{sec:method-1}

As shown in Fig.~\ref{fig:example}, our restoration pipeline is constructed upon the pretrained Stable Diffusion 2.1-base backbone. Given a hazy image $\mathbf{I}$,
the VAE encoder $\mathcal E$ produces a latent representation $\mathbf{z}_I$ and
intermediate features for subsequent reconstruction. At the fixed timestep
$t$$=$$399$, we use this encoded representation as the diffusion state, i.e.,
$\mathbf{z}_t$$=$$\mathbf{z}_I$. The standard forward
formulation is
\begin{equation}
    \mathbf{z}_t=\sqrt{\bar{\alpha}_t}\mathbf{z}+
    \sqrt{1-\bar{\alpha}_t}\boldsymbol{\epsilon},\qquad
    \bar{\alpha}_t=\prod_{s=1}^{t}(1-\beta_s),
    \label{eq:forward_diffusion}
\end{equation}
where $\mathbf{z}$ is a clean latent, $\boldsymbol{\epsilon}$ is Gaussian noise, and $\beta_s$
denotes the noise variance.

To account for spatially varying haze, SDA extracts multi-scale
degradation features directly from the hazy input. These features
guide multiple U-Net decoder stages and refine the output noise
prediction. The resulting prediction $\boldsymbol{\epsilon}'$ is used to recover the
clean latent in one denoising step:
\begin{equation}
    \widehat{\mathbf{z}}_0=
    \frac{\mathbf{z}_t-\sqrt{1-\bar{\alpha}_t}\boldsymbol{\epsilon}'}
    {\sqrt{\bar{\alpha}_t}}.
    \label{eq:one_step_denoising}
\end{equation}

The VAE decoder $\mathcal D$ then reconstructs the restored image from
$\widehat{\mathbf{z}}_0$. During decoding, ASC uses decoder features to query
encoder information. This supports detail reconstruction while reducing the influence of unreliable haze-degraded encoder features during image reconstruction.

For high-resolution inputs, we apply the same network to a downsampled
full image and overlapping patches at the original resolution. The
resulting global reference $\mathbf{G}$ guides the alignment and fusion of local
predictions $\{\mathbf{P}_i\}$. GPF further uses structural agreement with the hazy
input to compensate fine details, producing the final result $\widehat{\mathbf{I}}$.
Together, these designs enable spatially adaptive dehazing while
maintaining global appearance consistency and preserving local
structural details.

\subsection{Attentive Skip Compensation}
\label{sec:asc}

Latent diffusion models reduce computational cost through VAE compression,
but this may attenuate fine spatial details.
Encoder--decoder skip connections help recover these details by reusing
input features~\citep{wang2024exploiting,lan2025exploiting,ai2024multimodal,zhou2022towards,chen2025unirestore}.
However, spatially non-uniform haze causes the reliability of encoder
features to vary across regions.
Some regions retain useful scene details, whereas others are more
severely affected by haze.
Direct feature transfer can therefore introduce degraded information
along with useful details.
To address this variation, we propose attentive
skip compensation (ASC).
It selectively retrieves reliable encoder information and adaptively
refines less reliable features during reconstruction.

At stage $\ell$, ASC takes encoder and decoder features
$\mathbf{E}_\ell$ and $\mathbf{D}_\ell$. A lightweight convolutional
predictor with a sigmoid output estimates a spatial reliability map
$\mathbf{R}_\ell$ from both $\mathbf{E}_\ell$ and $\mathbf{D}_\ell$.
Decoder features then query encoder features through local cross-attention,
with reliability guiding the selection of source information. For each
window and attention head, the retrieved feature is
\begin{equation}
    \mathbf{V'}_\ell
    =
    \operatorname{softmax}\!\left(
        \frac{\mathbf{Q}_\ell \mathbf{K}_\ell^{\mathsf{T}}}{\sqrt{d}}
        + \mathbf{B}_\ell
        + \log\!\left(
            \max\{\mathbf{R}_\ell^{\mathsf{T}},
            \varepsilon\}
        \right)
    \right)\mathbf{V}_\ell,
    \label{eq:asc_retrieval}
\end{equation}
where $\mathbf{Q}_\ell$ and $(\mathbf{K}_\ell,\mathbf{V}_\ell)$
are projected from normalized decoder and encoder features, respectively.
$\mathbf{B}_\ell$ is the relative positional bias, $d$ is the head dimension, and $\varepsilon$$>$$0$ ensures
numerical stability.

Then we refine the encoder features using the difference between retrieved and source features, $\mathbf{V'}_\ell$$-$$\mathbf{V}_\ell$:
\begin{equation}
    \widehat{\mathbf{E}}_\ell
    =
    \mathbf{E}_\ell
    +
    (1-\mathbf{R}_\ell)\odot
    \operatorname{Conv}_{1\times1}
    \!\left(\mathbf{V'}_\ell-\mathbf{V}_\ell\right),
    \label{eq:asc_correction}
\end{equation}
where $\operatorname{Conv}_{1\times1}$ is a zero-initialized convolution
and $\odot$ denotes element-wise multiplication. The factor
$1$$-$$\mathbf{R}_\ell$ increases correction weights at less reliable
locations. The refined encoder features are concatenated with $\mathbf{D}_\ell$ and processed by residual convolutional blocks, whose output is added to $\mathbf{D}_\ell$.

This design retains reliable encoder information and focuses refinement
on less reliable regions. It supports detail reconstruction while
reducing the influence of haze-degraded features.

\subsection{Spatial Degradation Adapter}
\label{sec:sda}
Spatially non-uniform haze attenuates contrast and structural information
to varying degrees across an image, leading to different restoration needs.
To effectively handle spatially varying haze degradation, we introduce the spatial degradation adapter (SDA). It extracts multi-scale haze-aware features from the hazy input and injects them into the U-Net to guide spatially adaptive restoration.

\noindent \textbf{Haze representation.}
Given a hazy image $\mathbf{I}$, we use a convolutional encoder
to learn multi-scale haze representations
$\{\mathbf{F}_\ell\}_{\ell=1}^{3}$ that capture local degradation
and restoration needs. The supervision targets and training details
are provided in Appendix~\ref{app:training_sda_encoder}.

\begin{figure}[t]
    \centering
    \vspace{-2mm}
    \includegraphics[width=\linewidth]{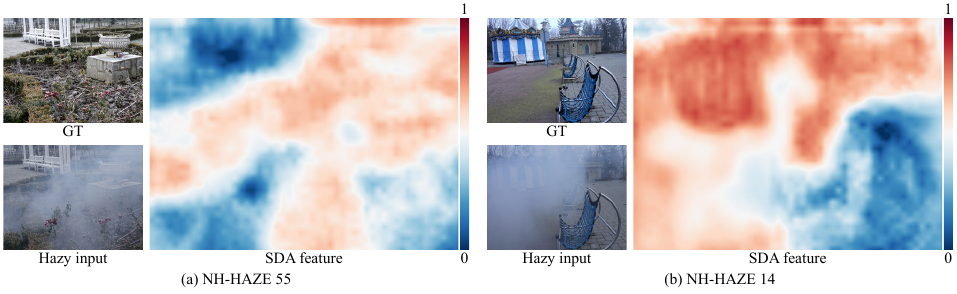}
    \caption{Visualization of the first SDA feature channel on NH-HAZE 55
    (left) and 14 (right). Both examples use the same color scale;
    warmer colors indicate stronger feature responses.}
    \label{fig:sda_features}
    \vspace{-2mm}
\end{figure}

To illustrate the haze information captured by SDA, we visualize the first
channel of its finest-scale features in Fig.~\ref{fig:sda_features}. On NH-HAZE 55,
the visible haze across the central region exhibits stronger responses, whereas the clearer
upper-left region generally shows weaker responses. This spatial
correspondence suggests that the learned features are sensitive to
variations in haze distribution. Additional SDA feature visualizations
on NH-HAZE and O-HAZE are provided in
Appendix~\ref{app:sda_feature_visualizations}.

\noindent \textbf{U-Net adaptation.}
We align the haze features with three U-Net decoder stages, where
convolutional adapters produce spatially gated residual corrections.
At each scale, we inject features as:
\begin{equation}
    \mathbf{F}_{\mathrm d}'=\mathbf{F}_{\mathrm d}
    +\operatorname{Adapter}(\mathbf{F}_{\mathrm d},\mathbf{F}),
    \label{eq:sda_modulation}
\end{equation}
where $\mathbf{F}$ denotes haze features,
$\mathbf{F}_{\mathrm d}$ and $\mathbf{F}_{\mathrm d}'$
are decoder features before and after injection, respectively,
and $\operatorname{Adapter}$ denotes the convolutional adapter. We also use the haze features to modulate the U-Net noise prediction, yielding $\boldsymbol{\epsilon}'$.
Both feature and noise correction branches use zero-initialized output
projections to preserve the original network output at initialization. Through multi-scale feature injection and noise prediction refinement,
SDA incorporates spatially varying haze information into one-step
denoising, adapting restoration to local degradation.

\subsection{Global-Guided Patch Fusion}
\label{sec:method-3}

\begin{figure*}[t]
    \centering
    \vspace{-3mm}
    \includegraphics[width=1.\textwidth]{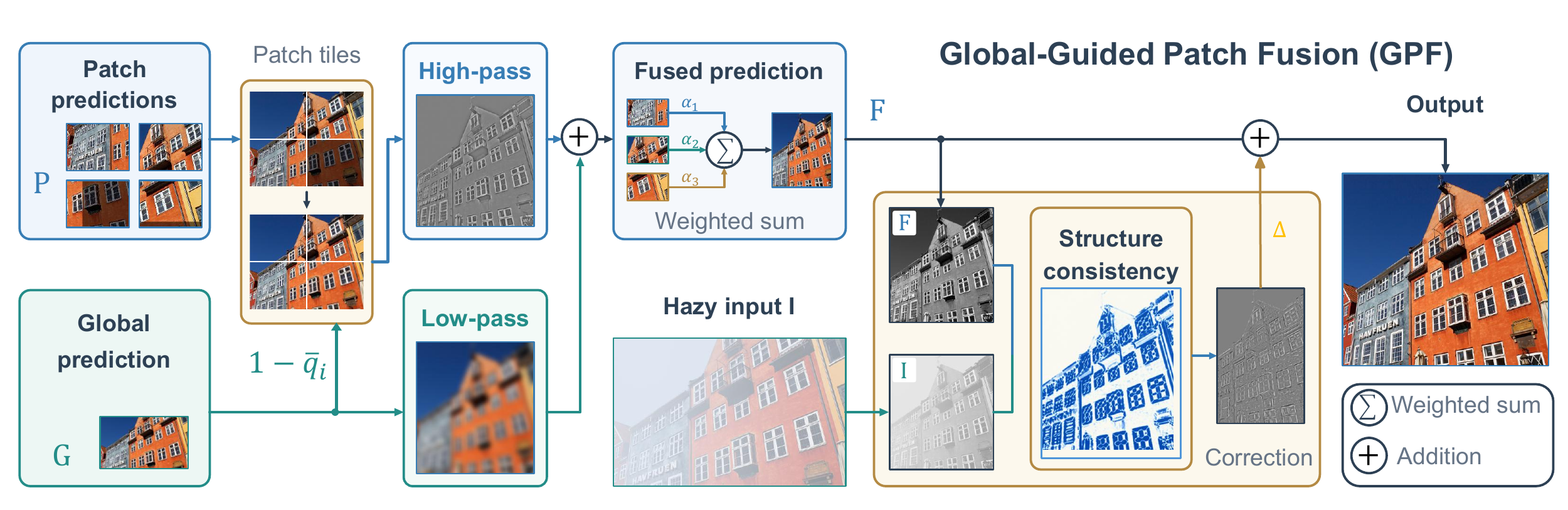}
    \caption{Overview of global-guided patch fusion (GPF).
    A global prediction guides patch alignment and reliability-aware fusion.
    Structural cues from the hazy input guide detail compensation.}
    \label{fig:postprocess}
\end{figure*}

High-resolution diffusion-based dehazing often relies on patch-wise inference
to accommodate limited GPU memory. However, each patch observes only partial
scene context, which can lead to inconsistent color and illumination across regions. Averaging overlapping predictions can alleviate boundary
artifacts, but cannot ensure global appearance consistency and may smooth fine
details.

To overcome these issues, we introduce global-guided patch fusion (GPF), a training-free strategy that combines global appearance guidance with local detail preservation.
As illustrated in Fig.~\ref{fig:postprocess}, a prediction from the downsampled full image provides a common appearance reference, while original-resolution
patch predictions supply complementary fine details. GPF consists of three components: appearance alignment, reliability-aware fusion, and detail compensation.

\noindent \textbf{Appearance alignment.}
For each local prediction $\mathbf{P}_i$, we combine its multiscale wavelet
high-frequency component with the low-frequency component of the corresponding
global region $\mathbf{G}_i$. This preserves local structures while establishing a
consistent global appearance.

\noindent \textbf{Reliability-aware fusion.}
Appearance alignment preserves local details by retaining the
high-frequency component of each patch prediction. However, patch predictions may contain artifacts due to limited scene context.
We therefore assess the reliability of each raw patch prediction
against the global reference and use it to control the contribution of the aligned prediction during fusion.

We downsample each raw patch prediction to the native resolution of the global branch.
We then compute its PSNR-Y against the corresponding global region.
A sigmoid mapping converts this score into a reliability weight $q_i$.
Consistent patches receive higher weights to preserve local details.
Inconsistent patches contribute less and rely more on the global reference.

To avoid abrupt transitions between patches, we aggregate their reliability weights
using cosine windows $\mathbf{w}_i$ and apply Gaussian smoothing $\mathcal S$.
The resulting spatial reliability map $\bar{\mathbf q}$ controls the blending
of aligned local predictions with the global reference:
\begin{equation}
    \bar{\mathbf{q}}=\mathcal S\!\left(\frac{\sum_i \mathbf{w}_iq_i}{\sum_i \mathbf{w}_i}\right),\quad
    \widehat{\mathbf{P}}_i=\bar{\mathbf{q}}_i\odot
    \left[\mathcal W_{\mathrm H}(\mathbf{P}_i)+\mathcal W_{\mathrm L}(\mathbf{G}_i)\right]
    +(1-\bar{\mathbf{q}}_i)\odot\mathbf{G}_i.
    \label{eq:global_guided_correction}
\end{equation}
$\mathbf{G}_i$ and $\bar{\mathbf{q}}_i$ denote the corresponding crops
of the global prediction and reliability map, respectively.
The operators $\mathcal W_{\mathrm H}$ and $\mathcal W_{\mathrm L}$ extract
the wavelet high- and low-frequency components, respectively.
Combining the two wavelet components yields the prediction produced by appearance alignment.
Reliable regions retain more local information, while unreliable regions rely
more on the global prediction.
Finally, we aggregate the corrected patches using reliability weights and cosine windows:
\begin{equation}
    \mathbf{I}^{\mathrm f}=\frac{\sum_i q_i\mathbf{w}_i\odot\widehat{\mathbf{P}}_i}{\sum_i q_i\mathbf{w}_i}.
    \label{eq:weighted_fusion}
\end{equation}

\noindent \textbf{Detail compensation.}
Aggregating patch predictions may attenuate high-frequency details.
We therefore use structural information retained in the hazy input to compensate
for detail loss in the fused result.
First, we estimate a local structural agreement map $\mathbf{s}$ between
the hazy input $\mathbf{I}$ and the fused result $\mathbf{I}^{\mathrm f}$.
Its values lie in $[0,1]$ and guide spatially adaptive detail compensation.

We then extract high-frequency information from both images using self-guided filtering,
obtaining detail layers $\mathbf{H}$ and $\mathbf{H}^{\mathrm f}$, respectively.
Guided by structural agreement, we adaptively retain restored details and incorporate
input details as follows:
\begin{equation}
    \widehat{\mathbf{H}}=\bigl[\alpha+\mathbf{s}(1-\alpha)\bigr]\odot\mathbf{H}^{\mathrm f}
    +\beta\mathbf{s}\odot\mathbf{H}.
    \label{eq:detail_compensation}
\end{equation}
The fixed coefficients $\alpha\in(0,1)$ and $\beta>0$ control the retention
of restored details and compensation from input details, respectively.
Regions with stronger structural agreement retain more restored details and receive
more detail from the input.
Regions with weaker agreement have attenuated detail responses, reducing the influence
of inconsistent structures.
Finally, we combine compensated details with low-frequency information from
the fused result to reconstruct the output $\widehat{\mathbf{I}}$.

Through appearance alignment, reliability-aware fusion,
and structure-guided detail compensation, GPF improves cross-patch consistency while preserving fine structures in high-resolution dehazing.

\begin{table}[t]
    \centering
    \caption{Dehazing performance of Hihaze and competing methods across four real-world benchmarks.
For each metric, \textcolor{red}{red} denotes first place and
\textcolor{blue}{blue} denotes second place.}
    \label{tab:comparison_main}
    \scriptsize
    \setlength{\tabcolsep}{2.890pt}
    \begin{tabular}{l|l|ccc|ccccc|c}
            \toprule[0.1em]
            \rowcolor{color-tab}
            & & \multicolumn{3}{c|}{Non-diffusion}
            & \multicolumn{6}{c}{Diffusion / Flow} \\
            \rowcolor{color-tab}
            \multirow{-2}{*}{Dataset} & \multirow{-2}{*}{Metric}
            & PTTD & IPC-Dehaze & BiLaLoRA & DiffUIR & FCDM & Diff-Dehazer & DiffDehaze & HazeFlow & Ours \\
            \midrule[0.1em]
            \multirow{6}{*}{NH-HAZE}
                & PSNR $\uparrow$ & \textcolor{blue}{14.53} & 11.18 & 12.07 & 11.79 & 14.04 & 12.99 & 14.26 & 14.38 & \textcolor{red}{18.22} \\
                & SSIM $\uparrow$ & 0.6119 & 0.5850 & 0.5989 & 0.4788 & 0.5955 & 0.6079 & 0.3067 & \textcolor{blue}{0.6343} & \textcolor{red}{0.6664} \\
                & LPIPS $\downarrow$ & \textcolor{blue}{0.3934} & 0.4445 & 0.4398 & 0.5424 & 0.4264 & 0.3936 & 0.6763 & 0.4180 & \textcolor{red}{0.2863} \\
                & DISTS $\downarrow$ & \textcolor{blue}{0.1938} & 0.2400 & 0.2106 & 0.3462 & 0.2159 & 0.2094 & 0.3374 & 0.2021 & \textcolor{red}{0.1272} \\
                & FID $\downarrow$ & 187.93 & 200.18 & 199.34 & 215.83 & 198.70 & \textcolor{blue}{151.51} & 170.95 & 162.45 & \textcolor{red}{90.10} \\
                & Q-Align $\uparrow$ & 3.4309 & 3.4721 & \textcolor{blue}{3.6929} & 3.1777 & 3.3738 & 3.4304 & 3.2871 & 3.1193 & \textcolor{red}{3.8066} \\
            \midrule
            \multirow{6}{*}{O-HAZE}
                & PSNR $\uparrow$ & \textcolor{blue}{18.78} & 13.98 & 12.75 & 15.12 & 15.35 & 16.62 & 15.77 & 16.63 & \textcolor{red}{21.28} \\
                & SSIM $\uparrow$ & 0.6241 & \textcolor{blue}{0.6601} & 0.6185 & 0.6413 & 0.5442 & 0.6332 & 0.5399 & 0.6484 & \textcolor{red}{0.6992} \\
                & LPIPS $\downarrow$ & \textcolor{blue}{0.3922} & 0.5383 & 0.4617 & 0.4558 & 0.4820 & 0.4085 & 0.3987 & 0.4291 & \textcolor{red}{0.3236} \\
                & DISTS $\downarrow$ & 0.1914 & 0.2406 & 0.1998 & 0.2975 & 0.2059 & 0.1935 & \textcolor{blue}{0.1763} & 0.2054 & \textcolor{red}{0.1087} \\
                & FID $\downarrow$ & 130.09 & 139.47 & 170.16 & 156.33 & 187.18 & 144.15 & \textcolor{blue}{124.64} & 152.20 & \textcolor{red}{66.86} \\
                & Q-Align $\uparrow$ & \textcolor{blue}{3.4973} & 3.3362 & 3.2017 & 2.9872 & 3.0500 & 3.2273 & 3.2613 & 3.3200 & \textcolor{red}{3.9219} \\
            \midrule
            \multirow{6}{*}{Dense-Haze}
                & PSNR $\uparrow$ & \textcolor{blue}{13.67} & 8.46 & 11.54 & 9.90 & 12.69 & 10.94 & 13.18 & 12.10 & \textcolor{red}{17.29} \\
                & SSIM $\uparrow$ & 0.4536 & 0.4548 & 0.5055 & 0.4432 & 0.4088 & 0.4656 & 0.4636 & \textcolor{blue}{0.5320} & \textcolor{red}{0.5676} \\
                & LPIPS $\downarrow$ & \textcolor{blue}{0.5865} & 0.7131 & 0.6492 & 0.7676 & 0.7813 & 0.7149 & 0.6786 & 0.6374 & \textcolor{red}{0.4774} \\
                & DISTS $\downarrow$ & \textcolor{blue}{0.3174} & 0.4617 & 0.4178 & 0.6137 & 0.4036 & 0.5034 & 0.4580 & 0.3834 & \textcolor{red}{0.2124} \\
                & FID $\downarrow$ & 256.00 & 271.59 & 262.50 & 272.00 & 275.20 & 256.24 & \textcolor{blue}{210.41} & 236.11 & \textcolor{red}{172.18} \\
                & Q-Align $\uparrow$ & 2.4576 & 2.2399 & \textcolor{blue}{2.9652} & 1.9165 & 2.0283 & 2.1314 & 2.5013 & 2.4468 & \textcolor{red}{3.3325} \\
            \midrule
            \multirow{6}{*}{I-HAZE}
                & PSNR $\uparrow$ & 15.84 & 15.68 & 15.67 & 15.89 & 14.09 & 15.83 & 13.89 & \textcolor{blue}{16.12} & \textcolor{red}{19.02} \\
                & SSIM $\uparrow$ & 0.6508 & \textcolor{red}{0.7985} & 0.7075 & 0.7705 & 0.4737 & 0.6709 & 0.7138 & 0.7675 & \textcolor{blue}{0.7888} \\
                & LPIPS $\downarrow$ & 0.5072 & 0.4595 & 0.4638 & \textcolor{blue}{0.3336} & 0.6258 & 0.4421 & 0.3725 & 0.4559 & \textcolor{red}{0.3217} \\
                & DISTS $\downarrow$ & 0.2172 & 0.2033 & \textcolor{blue}{0.1566} & 0.2448 & 0.2370 & 0.1849 & 0.1861 & 0.1974 & \textcolor{red}{0.0946} \\
                & FID $\downarrow$ & 197.12 & \textcolor{blue}{118.08} & 217.57 & 120.02 & 254.86 & 157.31 & 153.27 & 168.12 & \textcolor{red}{78.99} \\
                & Q-Align $\uparrow$ & 3.3117 & 3.4398 & \textcolor{blue}{3.6040} & 2.8698 & 2.9932 & 3.1771 & 3.3588 & 3.4755 & \textcolor{red}{4.2306} \\
            \bottomrule[0.1em]
    \end{tabular}
\end{table}
\section{Experiments}

\subsection{Experimental Settings}
\label{sec:experimental_settings}
\noindent \textbf{Datasets.}
Model training uses paired hazy and clean images from LMHaze~\citep{zhang2024lmhaze} and RW2AH~\citep{fang2025guided}. For evaluation, we adopt four real-world benchmarks covering diverse degradation conditions: indoor scenes in I-HAZE~\citep{ancuti2018ihaze}, outdoor environments in O-HAZE~\citep{ancuti2018haze}, heavy homogeneous haze in Dense-Haze~\citep{ancuti2019dense}, and spatially varying haze in NH-HAZE~\citep{ancuti2020nh}. We evaluate our model and other methods on all four datasets without fine-tuning on the target datasets.

\noindent \textbf{Evaluation Metrics.}
We evaluate reconstruction fidelity with PSNR and SSIM~\citep{wang2004image}, and perceptual agreement with LPIPS~\citep{zhang2018unreasonable} and DISTS~\citep{ding2020image}. FID~\citep{heusel2017gans} measures distributional differences
in feature space between restored images and clean references. Q-Align~\citep{wu2023q} provides no-reference quality scores.

\noindent \textbf{Implementation Details.}
We initialize the network from Stable Diffusion 2.1-base~\citep{rombach2022high}
and use LoRA with rank 16. With AdamW~\citep{loshchilov2017fixing},
we train the U-Net LoRA and ASC for 200,000 iterations on $512$$\times$$512$
crops. We then train the SDA encoder on $512$$\times$$512$ crops and the
U-Net adaptation branches on $1{,}024\times1{,}024$ crops for 10,000 iterations
each. Inference uses one denoising step at $t$$=$$399$; high-resolution
images are processed with overlapping $1{,}024\times1{,}024$ patches and GPF.
Appendix~\ref{app:training_details} in the supplementary material
documents the training procedure, losses, learning rates,
batch sizes, and other settings.
\subsection{Comparison with State-of-the-Art Methods}
\label{sec:comparison}
\noindent \textbf{Compared Methods.}
We compare our method with eight representative dehazing approaches.
The non-diffusion methods include PTTD~\citep{chen2024prompt},
IPC-Dehaze~\citep{fu2025iterative}, and
BiLaLoRA~\citep{zhang2026bilevel}.
The diffusion-based methods include
DiffUIR~\citep{zheng2024selective},
FCDM~\citep{wang2024frequency},
Diff-Dehazer~\citep{lan2025exploiting},
DiffDehaze~\citep{wang2025learning}, and
HazeFlow~\citep{shin2025hazeflow}.
\textbf{Appendix~\ref{app:additional_comparisons} extends the comparison to five
additional methods, reporting all six metrics on the same four benchmarks.}

\noindent \textbf{Quantitative Results.}
Across the four benchmarks in Tab.~\ref{tab:comparison_main}, Hihaze attains the highest PSNR and Q-Align scores and the lowest LPIPS, DISTS, and FID values. In particular, it outperforms the runner-up by 3.62 dB on Dense-Haze and 3.69 dB on NH-HAZE in PSNR, while reducing FID by 57.78 on O-HAZE and 61.41 on NH-HAZE. It also ranks first in SSIM on three benchmarks. These results support reconstruction fidelity and perceptual quality in real-world dehazing.

\begin{figure}[t]
    \centering

    \begingroup
    \scriptsize
    \setlength{\tabcolsep}{0.000517063\linewidth}
    \renewcommand{\arraystretch}{0}
    \begin{tabular}{@{}cccccccc@{}}
    \includegraphics[width=0.124095140\linewidth]{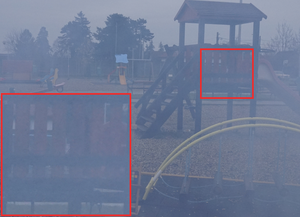} &
    \includegraphics[width=0.124095140\linewidth]{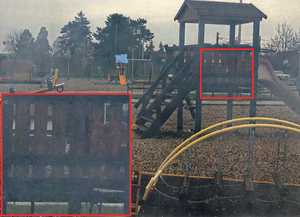} &
    \includegraphics[width=0.124095140\linewidth]{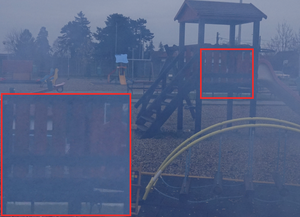} &
    \includegraphics[width=0.124095140\linewidth]{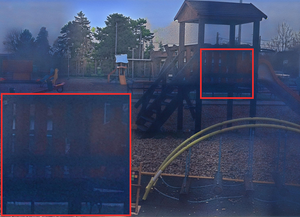} &
    \includegraphics[width=0.124095140\linewidth]{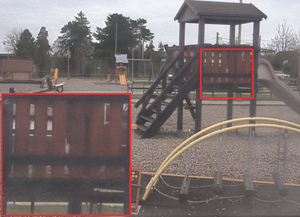} &
    \includegraphics[width=0.124095140\linewidth]{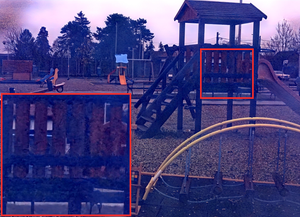} &
    \includegraphics[width=0.124095140\linewidth]{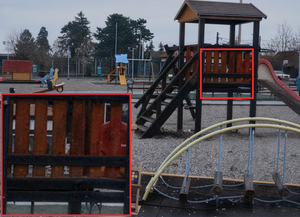} &
    \includegraphics[width=0.124095140\linewidth]{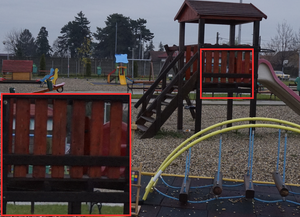} \\[0.273bp]
    \includegraphics[width=0.124095140\linewidth]{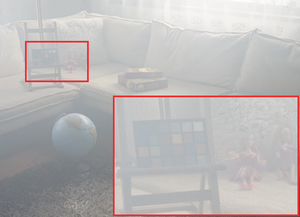} &
    \includegraphics[width=0.124095140\linewidth]{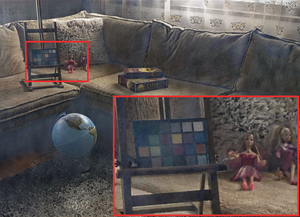} &
    \includegraphics[width=0.124095140\linewidth]{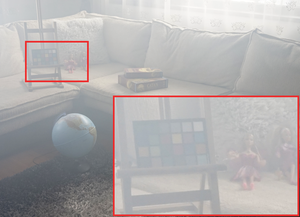} &
    \includegraphics[width=0.124095140\linewidth]{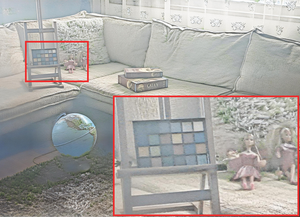} &
    \includegraphics[width=0.124095140\linewidth]{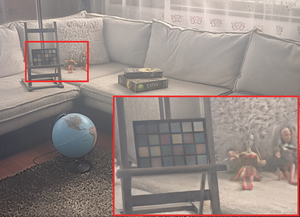} &
    \includegraphics[width=0.124095140\linewidth]{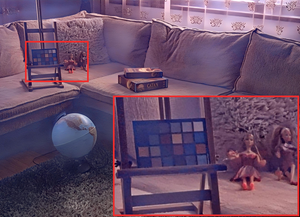} &
    \includegraphics[width=0.124095140\linewidth]{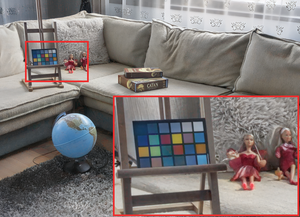} &
    \includegraphics[width=0.124095140\linewidth]{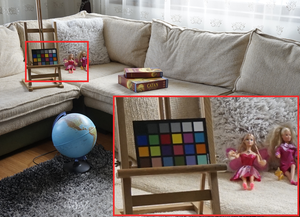} \\[0.273bp]
    \includegraphics[width=0.124095140\linewidth]{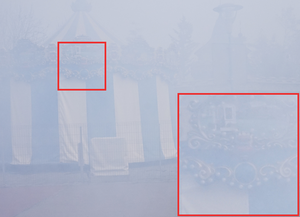} &
    \includegraphics[width=0.124095140\linewidth]{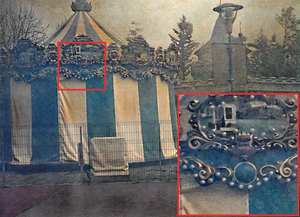} &
    \includegraphics[width=0.124095140\linewidth]{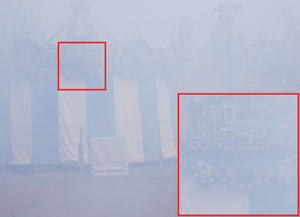} &
    \includegraphics[width=0.124095140\linewidth]{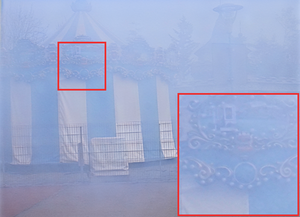} &
    \includegraphics[width=0.124095140\linewidth]{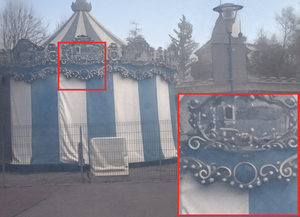} &
    \includegraphics[width=0.124095140\linewidth]{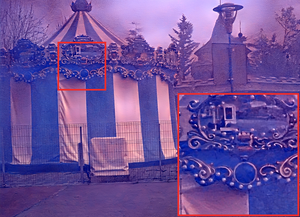} &
    \includegraphics[width=0.124095140\linewidth]{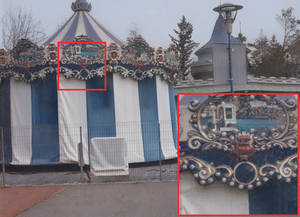} &
    \includegraphics[width=0.124095140\linewidth]{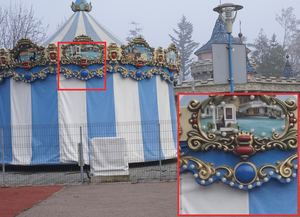} \\[0.273bp]
    \includegraphics[width=0.124095140\linewidth]{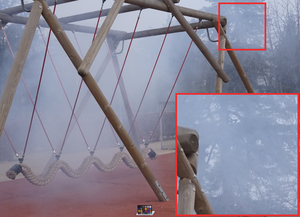} &
    \includegraphics[width=0.124095140\linewidth]{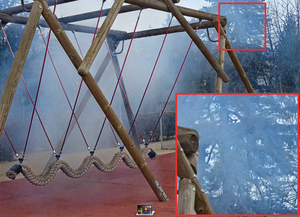} &
    \includegraphics[width=0.124095140\linewidth]{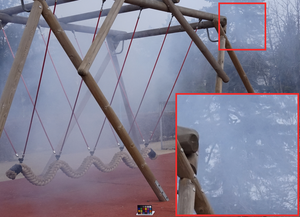} &
    \includegraphics[width=0.124095140\linewidth]{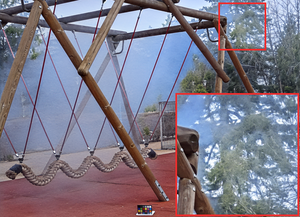} &
    \includegraphics[width=0.124095140\linewidth]{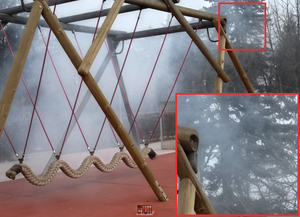} &
    \includegraphics[width=0.124095140\linewidth]{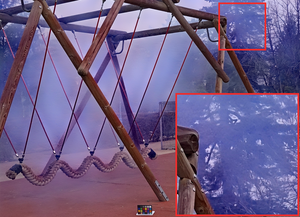} &
    \includegraphics[width=0.124095140\linewidth]{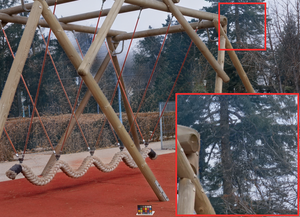} &
    \includegraphics[width=0.124095140\linewidth]{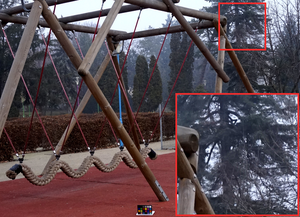} \\
    \rule{0pt}{9pt}Hazy input & PTTD & DiffUIR & Diff-Dehazer & DiffDehaze & HazeFlow & \textbf{Ours} & GT \\
    \end{tabular}
    \endgroup
    \caption{Visual comparisons on O-HAZE, I-HAZE,
    Dense-Haze, and NH-HAZE, from top to bottom. Red boxes mark the corresponding regions shown in the enlarged views.}
\label{fig:qualitative_comparison}
\end{figure}

\noindent \textbf{Qualitative Results.}
We provide visual comparisons on I-HAZE, O-HAZE, Dense-Haze, and NH-HAZE
in Fig.~\ref{fig:qualitative_comparison}. Our method produces more natural and haze-free results across diverse real-world scenes. Compared with
existing methods, it more effectively suppresses residual haze while
preserving structural and textural details under spatially non-uniform haze.
Dense-Haze remains particularly challenging: its extremely dense haze
severely obscures scene appearance, and our restored result still exhibits
some color discrepancies relative to the ground truth. The four examples
show scene views and enlarged local details; \textbf{more comparisons appear in Appendix~\ref{app:full_image_comparisons}.}

\begin{table}[t]
\centering
\caption{Inference time and model complexity for 3,328$\times$4,096 inputs on an RTX A6000.}
    \scriptsize
    \label{tab:efficiency}
    \setlength{\tabcolsep}{9.65pt}
    \begin{tabular}{l|cccccc|c}
            \toprule[0.1em]
            \rowcolor{color-tab}
            Metric & IPC-Dehaze & DiffUIR & FCDM & Diff-Dehazer & DiffDehaze & HazeFlow & Ours \\
            \midrule[0.1em]
            Runtime (s) & 185.4 & 43.5 & 146.6 & 32.5 & 1,046.8 & 15.2 & 12.9 \\
            FLOPs (T) & 826.391 & 236.731 & 2,022.874 & 453.871 & 9,386.778 & 250.991 & 458.955 \\
            Params (M) & 43.164 & 36.261 & 155.334 & 1,506.159 & 1,666.750 & 87.087 & 975.822 \\
            \bottomrule[0.1em]
    \end{tabular}
    
\end{table}

\noindent \textbf{Running Time Comparisons.}
Tab.~\ref{tab:efficiency} summarizes inference latency, FLOPs, and parameter counts. All methods are tested
on the same NVIDIA RTX A6000 GPU with $3{,}328\times4{,}096$ input
images. Our method takes 12.9~s per image, achieving approximately
an $81\times$ speedup over DiffDehaze. Even compared with HazeFlow,
the fastest competing method, our method reduces the running time
from 15.2~s to 12.9~s, achieving a $1.18\times$ speedup.

\subsection{Ablation Studies}
\label{sec:ablation}
We investigate the effectiveness of attentive skip compensation (ASC),
spatial degradation adapter (SDA), and global-guided patch fusion (GPF)
on O-HAZE. Specifically, we compare different VAE skip-connection designs
and conditioning strategies, and progressively introduce the GPF
components to examine their individual contributions. We further
evaluate the transferability of GPF to two other dehazing methods
without retraining. All variants are evaluated using PSNR, SSIM, LPIPS, and DISTS
to assess fidelity and perceptual similarity, as shown in
Tab.~\ref{tab:ablation}.

\begin{table}[t]
    \centering
    \setlength{\parskip}{0pt}
    \setlength{\tabcolsep}{1.5pt}
    \caption{Ablation studies and GPF transferability on O-HAZE. (a)  Base denotes the model without skip connections; Zero-conv uses zero-conv skip connections. (b) Base denotes the model with ASC. (c) AA: appearance alignment; RF: reliability-aware fusion;
DC: detail compensation. (d) Performance comparisons of HazeFlow and Diff-Dehazer
with and without GPF.}
    \label{tab:ablation}
    \subfloat[Ablation of ASC.\label{tab:ablation_skip}]{%
    \begin{minipage}[t]{0.49\linewidth}
    \centering
    \scriptsize
    \setlength{\tabcolsep}{12pt}
    \begin{tabular}{l|ccc}
            \toprule
            \rowcolor{color-tab}
            Metric & base & + Zero-conv & + ASC \\
            \midrule
            PSNR $\uparrow$ & 17.36 & 18.80 & \textbf{19.79} \\
            SSIM $\uparrow$ & 0.5636 & 0.6438 & \textbf{0.6508} \\
            LPIPS $\downarrow$ & 0.3837 & 0.3484 & \textbf{0.3418} \\
            DISTS $\downarrow$ & 0.1465 & \textbf{0.1327} & 0.1334 \\
            \bottomrule
    \end{tabular}
    \end{minipage}}\hfill%
    \subfloat[Ablation of SDA.\label{tab:ablation_vdp}]{%
    \begin{minipage}[t]{0.49\linewidth}
    \centering
    \scriptsize
    \setlength{\tabcolsep}{4.2635pt}
    \begin{tabular}{l|cccc}
            \toprule
            \rowcolor{color-tab}
            Method & PSNR $\uparrow$ & SSIM $\uparrow$ & LPIPS $\downarrow$
            & DISTS $\downarrow$ \\
            \midrule
            base & 19.79 & 0.6508 & 0.3418 & 0.1334 \\
            + T2I-Adapter & 19.78 & 0.6500 & 0.3404 & 0.1235 \\
            + Noise refinement & 19.78 & 0.6483 & 0.3406 & 0.1291 \\
            + SDA & \textbf{20.08} & \textbf{0.6598} & \textbf{0.3375} & \textbf{0.1216} \\
            \bottomrule
    \end{tabular}
    \end{minipage}}\par\vspace{0pt}
    \subfloat[Ablation of GPF.\label{tab:ablation_postprocess}]{%
    \begin{minipage}[t]{0.49\linewidth}
    \centering
    \scriptsize
    \renewcommand{\arraystretch}{1.1255}
    \setlength{\tabcolsep}{5.0723pt}
    \begin{tabular}{ccc|cccc}
            \toprule
            \rowcolor{color-tab}
            AA & RF & DC & PSNR $\uparrow$ & SSIM $\uparrow$ & LPIPS $\downarrow$
            & DISTS $\downarrow$ \\
            \midrule
            -- & -- & -- & 20.08 & 0.6598 & 0.3375 & 0.1216 \\
            $\checkmark$ & -- & -- & 21.08 & 0.6664 & 0.3282 & 0.1166 \\
            $\checkmark$ & $\checkmark$ & -- & 21.27 & 0.6979 & 0.3285 & 0.1101 \\
            $\checkmark$ & $\checkmark$ & $\checkmark$ & \textbf{21.28} & \textbf{0.6992} & \textbf{0.3236} & \textbf{0.1087} \\
            \bottomrule
    \end{tabular}
    \end{minipage}}\hfill%
    \subfloat[GPF transferability.\label{tab:gpf_transfer_ohaze}]{%
    \begin{minipage}[t]{0.49\linewidth}
    \centering
    \scriptsize
    \setlength{\tabcolsep}{6.2270pt}
    \begin{tabular}{l|cccc}
            \toprule
            \rowcolor{color-tab}
            Method & PSNR $\uparrow$ & SSIM $\uparrow$ & LPIPS $\downarrow$
            & DISTS $\downarrow$ \\
            \midrule
            HazeFlow & 16.63 & 0.6484 & \textbf{0.4291} & 0.2054 \\
            \quad + GPF & \textbf{17.95} & \textbf{0.6782} & 0.4295 & \textbf{0.1732} \\
            \midrule
            Diff-Dehazer & 16.62 & 0.6332 & \textbf{0.4085} & 0.1935 \\
            \quad + GPF & \textbf{16.78} & \textbf{0.6456} & 0.4089 & \textbf{0.1783} \\
            \bottomrule
    \end{tabular}
    \end{minipage}}
    \vspace{-6mm}
\end{table}

\noindent \textbf{Attentive skip compensation.}
We compare three VAE variants in Tab.~\ref{tab:ablation_skip}: without
skip connections, with zero-conv skip connections, and with ASC.
Adding zero-conv improves all four metrics over the no-skip baseline,
showing the benefit of introducing encoder features into reconstruction.
ASC further improves PSNR, SSIM, and LPIPS by adaptively compensating
encoder features, while zero-conv achieves the lowest DISTS.
These results support the effectiveness of ASC for reconstruction.

\noindent \textbf{Spatial degradation adapter.}
We compare different conditioning strategies using the model with ASC
as the baseline. As shown in Tab.~\ref{tab:ablation_vdp}, T2I-Adapter
and noise refinement improve LPIPS and DISTS, but slightly reduce PSNR
and SSIM. By incorporating spatially varying haze information, SDA
improves all four metrics over the baseline and outperforms the alternative conditioning designs. This suggests that haze-aware
conditioning provides effective guidance for restoration.

\noindent \textbf{Global-guided patch fusion.}
We progressively add the GPF components in Tab.~\ref{tab:ablation_postprocess}.
First, appearance alignment improves all four metrics by aligning local
predictions with the global reference. Then, reliability-aware fusion
weights patches according to their consistency with the global prediction,
further improving PSNR, SSIM, and DISTS. Finally, detail compensation
uses structural cues from the hazy input and improves all four metrics,
supporting the contribution of each component.

\noindent \textbf{GPF transferability.}
We apply the same GPF procedure to HazeFlow~\citep{shin2025hazeflow}
and Diff-Dehazer~\citep{lan2025exploiting}.
As shown in Tab.~\ref{tab:gpf_transfer_ohaze}, GPF improves PSNR,
SSIM, and DISTS for both methods on O-HAZE, although LPIPS increases
slightly. These results indicate that the benefits of GPF extend to
other dehazing methods. Appendix~\ref{app:gpf_transfer} provides
results for all six metrics on all four benchmarks, together with
qualitative comparisons before and after applying GPF to further
examine its effectiveness across different methods and haze conditions.

\vspace{-2mm}
\section{Conclusion}
\vspace{-2mm}

We have presented Hihaze, a single-step latent diffusion approach tailored for restoring high-resolution hazy scenes. To handle spatially non-uniform haze, Hihaze employs ASC to preserve reliable encoder features and SDA to inject spatial degradation information into one-step dehazing. Moreover, GPF uses a global prediction to guide local patch fusion, improving cross-patch consistency while preserving fine details. Extensive experiments on real-world benchmarks demonstrate superior restoration quality and inference efficiency across diverse haze conditions.

\subsection*{AI Use Statement}
We utilized generative AI solely for manuscript language refinement, such as grammatical checking, phrasing adjustments, and readability enhancement. Every section edited with AI assistance has been thoroughly verified by the authors, who assume complete responsibility for the manuscript's accuracy and integrity. All experimental data generation, evaluation, and result analyses were performed entirely by the authors without any involvement of generative AI.

\subsection*{Ethics Statement}
This work studies single-image dehazing using existing research datasets
and pretrained models. No human-subject studies were conducted. We do not identify any specific ethical concerns arising from this research on restoring hazy images using existing data.

\subsection*{Reproducibility Statement}
Section~\ref{sec:experimental_settings} describes the training datasets,
evaluation benchmarks, and implementation settings.
Appendix~\ref{app:training_details} of the supplementary material provides
the complete training strategy, loss functions, and detailed configurations.
We plan to release code and trained weights upon acceptance.

\bibliography{conf_bib}
\bibliographystyle{iclr2027_conference}

\appendix
\begingroup
\section{Training Strategy and Implementation Details}
\label{app:training_details}
We train Hihaze in two stages.
In Stage~1, we jointly train the restoration backbone and ASC.
In Stage~2, we freeze the base restoration model and sequentially
train the SDA encoder and adaptation branches.
GPF requires no training.

\subsection{Backbone and ASC Training}
\label{app:training_backbone}

We initialize the model from Stable Diffusion 2.1-base and train
the U-Net LoRA parameters, ASC and its feature-fusion layers,
and a shared learnable text embedding.
The remaining pretrained parameters are kept frozen.
The LoRA rank and scaling parameter are both set to 16.

The restoration loss is defined as
\begin{equation}
    \mathcal{L}_{\mathrm b}
    =
    \mathcal{L}_1
    + \lambda_{\mathrm{pal}}\mathcal{L}_{\mathrm{pal}}
    + \lambda_{\mathrm{per}}\mathcal{L}_{\mathrm{per}}
    + \lambda_{\mathrm{adv}}\mathcal{L}_{\mathrm{adv}},
    \label{eq:restoration_loss}
\end{equation}
where $\mathcal{L}_1$ is the L1 reconstruction loss,
$\mathcal{L}_{\mathrm{pal}}$ is the affinely aligned pixel
loss (PAL)~\citep{li2026global}, and
$\mathcal{L}_{\mathrm{per}}$ is the sum of DISTS losses
on RGB images and their Scharr edge-magnitude
maps~\citep{li2026unleashing}.
We set $\lambda_{\mathrm{pal}}=0.8$,
$\lambda_{\mathrm{per}}=1$, and
$\lambda_{\mathrm{adv}}=0.1$.
PAL uses an affine RGB transformation with a bias term and
a ridge regularization coefficient of $10^{-3}$,
with gradients stopped through the fitted transformation.
The adversarial loss is computed using a DINOv3-based
discriminator~\citep{simeoni2025dinov3}.
Its ConvNeXt-Large feature extractor is frozen, while its
head is updated alternately with the restoration model.

ASC reliability supervision is constructed from paired hazy
and clean images $\mathbf{I},\mathbf{I}_{\mathrm c}\in[0,1]$:
\begin{equation}
    \mathbf{R}^{*}
    =
    \exp\!\left[
        -\alpha_{\mathrm{rel}}\,
        \operatorname{Avg}_{11\times11}\!\left(
            \operatorname{Mean}_{c}
            \left|\mathbf{I}-\mathbf{I}_{\mathrm c}\right|
        \right)
    \right],
    \label{eq:asc_reliability_target}
\end{equation}
where $\operatorname{Mean}_{c}$ and
$\operatorname{Avg}_{11\times11}$ denote channel averaging
and local spatial averaging, respectively, and
$\alpha_{\mathrm{rel}}=4$.
We average-pool the target map to each ASC resolution and
supervise the predicted reliability maps with an L1 loss
weighted by $0.05$.
The inputs to the reliability predictor and the reliability
values used for attention routing and gating are detached.

\subsection{SDA Training}
\label{app:training_sda}

\subsubsection{Encoder Training}
\label{app:training_sda_encoder}

Given a hazy input $\mathbf{I}$, its clean reference
$\mathbf{I}_{\mathrm c}$, and the output
$\mathbf{I}_{\mathrm b}$ of the frozen base model,
we construct three supervision targets for structural
attenuation, residual correction, and structural confidence:
\begin{equation}
    \mathbf{T}_{\mathrm d}
    = 1-\frac{\Phi(\mathbf{I})+\varepsilon_{\mathrm{str}}}
    {\Phi(\mathbf{I}_{\mathrm c})+\varepsilon_{\mathrm{str}}},
    \quad
    \mathbf{T}_{\mathrm r}
    = \frac{\Phi(\mathbf{I}_{\mathrm c})-\Phi(\mathbf{I}_{\mathrm b})}
    {\Phi(\mathbf{I}_{\mathrm c})+\varepsilon_{\mathrm{str}}},
    \quad
    \mathbf{T}_{\mathrm s}
    = \operatorname{sigmoid}\!\left(
        \frac{\Phi(\mathbf{I}_{\mathrm c})-\tau}{\kappa}
    \right).
    \label{eq:sda_targets}
\end{equation}
Here, $\Phi(\cdot)$ combines multiscale local Sobel-gradient
and contrast statistics.
We set $\varepsilon_{\mathrm{str}}=10^{-3}$,
$\tau=0.03$, and $\kappa=0.01$, and clip
$\mathbf{T}_{\mathrm d}$ and $\mathbf{T}_{\mathrm r}$
to $[0,1]$ and $[-1,1]$, respectively.
These targets supervise the multiscale SDA features,
and the encoder is trained using a Smooth L1 loss.

\subsubsection{Adaptation Branch Training}
\label{app:training_sda_adaptation}

We then freeze the SDA encoder and train only the three
adaptation branches in the U-Net decoder and the
noise-correction branch.
Their output projections are zero-initialized.
Training uses the same restoration loss and weights as
in Stage~1, with an additional mean squared penalty on
the noise correction weighted by $10^{-5}$.
The entire discriminator remains frozen.
The L1, PAL, and adversarial losses are computed on
$1{,}024\times1{,}024$ crops, while the RGB and edge
DISTS losses are evaluated after resizing the images
to $512\times512$.

\subsection{Implementation Details}
\label{app:training_common}

All training phases use AdamW, with configurations summarized
in Table~\ref{tab:training_configuration}.
Each optimizer uses 500 warmup updates, followed by the
learning-rate schedule specified in the table.

\begin{table}[t]
    \centering
    \caption{Training configurations for Hihaze.}
    \label{tab:training_configuration}
    \small
    \setlength{\tabcolsep}{4pt}
    \renewcommand{\arraystretch}{1.15}
    \resizebox{\linewidth}{!}{
        \begin{tabular}{lccc}
            \hline
            Configuration
            & Backbone and ASC
            & SDA encoder
            & SDA adaptation branches\\
            \hline
            Training iterations
            & $200{,}000$
            & $10{,}000$
            & $10{,}000$\\

            Crop size
            & $512\times512$
            & $512\times512$
            & $1{,}024\times1{,}024$\\

            Total batch size
            & 4
            & 1
            & 4\\

            Learning rate
            & $5\times10^{-5}$
            & $2\times10^{-4}$
            & $10^{-4}$\\

            Weight decay
            & $0.01$
            & $10^{-4}$
            & $0.01$\\

            Learning-rate schedule
            & Constant
            & Cosine to $10\%$ of peak
            & Cosine to zero\\
            \hline
        \end{tabular}
    }
\end{table}

In Stage~1, the discriminator head uses a learning rate of
$10^{-5}$ and a weight decay of $0.01$, with a constant
learning rate after warmup.
Inputs to the SDA encoder are resized to a long side of 512
while preserving their aspect ratio.

Paired images share the same cropping and augmentation.
Backbone training uses random flips and rotations, while
SDA training uses horizontal and vertical flips.
Images smaller than the required crop size are resized
while preserving their aspect ratio before cropping.

\section{Additional Quantitative Comparisons}
\label{app:additional_comparisons}

Table~\ref{tab:comparison_additional} reports results for the five methods
omitted from the compact comparison in the main paper:
DehazeFormer~\citep{song2022vision}, RIDCP~\citep{wu2023ridcp},
CORUN~\citep{fang2024real}, CoA~\citep{ma2025coa}, and
DehazeSB~\citep{lan2025schrodinger}. The table uses the same four benchmarks
and six metrics as Sec.~\ref{sec:comparison}, and repeats our results to
allow direct comparison. Red and blue identify the best and second-best
values within this table, respectively.

Our method achieves the best values for all six metrics on each benchmark
among the methods in Table~\ref{tab:comparison_additional}. Together with
Table~\ref{tab:comparison_main}, these results extend the comparison beyond
the methods presented in the main paper, covering both non-diffusion and
diffusion-based approaches.

\begin{table}[!t]
    \centering
    \caption{Additional quantitative comparisons on I-HAZE, O-HAZE,
    Dense-Haze, and NH-HAZE. The best and second-best values within this
    table are colored in \textcolor{red}{red} and \textcolor{blue}{blue}, respectively.}
    \label{tab:comparison_additional}
    \scriptsize
    \setlength{\tabcolsep}{8.550pt}
    \begin{tabular}{l|l|ccccc|c}
            \toprule[0.1em]
            \rowcolor{color-tab}
            & & \multicolumn{5}{c|}{Non-diffusion}
            & Diffusion / Flow \\
            \rowcolor{color-tab}
            \multirow{-2}{*}{Dataset} & \multirow{-2}{*}{Metric}
            & DehazeFormer & RIDCP & CORUN & CoA & DehazeSB & Ours \\
            \midrule[0.1em]
            \multirow{6}{*}{I-HAZE}
                & PSNR $\uparrow$ & 16.38 & 15.63 & \textcolor{blue}{17.10} & 13.82 & 16.04 & \textcolor{red}{19.02} \\
                & SSIM $\uparrow$ & \textcolor{blue}{0.7835} & 0.7116 & 0.7658 & 0.4888 & 0.6135 & \textcolor{red}{0.7888} \\
                & LPIPS $\downarrow$ & \textcolor{blue}{0.3360} & 0.4158 & 0.4093 & 0.6303 & 0.5019 & \textcolor{red}{0.3217} \\
                & DISTS $\downarrow$ & 0.2188 & \textcolor{blue}{0.1645} & 0.2142 & 0.2512 & 0.1783 & \textcolor{red}{0.0946} \\
                & FID $\downarrow$ & \textcolor{blue}{135.46} & 204.88 & 136.26 & 214.98 & 202.65 & \textcolor{red}{78.99} \\
                & Q-Align $\uparrow$ & 2.9887 & 3.6969 & 3.5317 & \textcolor{blue}{3.7021} & 3.3250 & \textcolor{red}{4.2306} \\
            \midrule
            \multirow{6}{*}{O-HAZE}
                & PSNR $\uparrow$ & \textcolor{blue}{16.36} & 14.14 & 15.02 & 9.99 & 15.20 & \textcolor{red}{21.28} \\
                & SSIM $\uparrow$ & \textcolor{blue}{0.6657} & 0.6244 & 0.6179 & 0.5851 & 0.6346 & \textcolor{red}{0.6992} \\
                & LPIPS $\downarrow$ & 0.4431 & \textcolor{blue}{0.4166} & 0.5104 & 0.5375 & 0.4569 & \textcolor{red}{0.3236} \\
                & DISTS $\downarrow$ & 0.2734 & 0.2008 & 0.2729 & 0.2492 & \textcolor{blue}{0.1910} & \textcolor{red}{0.1087} \\
                & FID $\downarrow$ & 153.79 & 175.76 & \textcolor{blue}{148.36} & 206.91 & 161.83 & \textcolor{red}{66.86} \\
                & Q-Align $\uparrow$ & 3.0723 & \textcolor{blue}{3.5222} & 3.3406 & 3.3174 & 3.4124 & \textcolor{red}{3.9219} \\
            \midrule
            \multirow{6}{*}{Dense-Haze}
                & PSNR $\uparrow$ & 10.85 & 9.95 & 10.44 & 9.59 & \textcolor{blue}{11.11} & \textcolor{red}{17.29} \\
                & SSIM $\uparrow$ & 0.4621 & 0.4686 & \textcolor{blue}{0.4911} & 0.4874 & 0.4210 & \textcolor{red}{0.5676} \\
                & LPIPS $\downarrow$ & 0.7456 & 0.6827 & 0.6858 & 0.7166 & \textcolor{blue}{0.6686} & \textcolor{red}{0.4774} \\
                & DISTS $\downarrow$ & 0.6006 & 0.4515 & 0.4090 & 0.4161 & \textcolor{blue}{0.3499} & \textcolor{red}{0.2124} \\
                & FID $\downarrow$ & 270.10 & 294.78 & \textcolor{blue}{251.70} & 265.38 & 277.73 & \textcolor{red}{172.18} \\
                & Q-Align $\uparrow$ & 1.9896 & 2.6232 & 2.5245 & 2.6543 & \textcolor{blue}{2.9389} & \textcolor{red}{3.3325} \\
            \midrule
            \multirow{6}{*}{NH-HAZE}
                & PSNR $\uparrow$ & 12.29 & 12.50 & 12.43 & 11.04 & \textcolor{blue}{12.69} & \textcolor{red}{18.22} \\
                & SSIM $\uparrow$ & 0.4959 & 0.5416 & 0.5738 & 0.5731 & \textcolor{blue}{0.5907} & \textcolor{red}{0.6664} \\
                & LPIPS $\downarrow$ & 0.5300 & 0.4479 & 0.5664 & 0.4986 & \textcolor{blue}{0.4421} & \textcolor{red}{0.2863} \\
                & DISTS $\downarrow$ & 0.3258 & 0.2306 & 0.2735 & 0.2262 & \textcolor{blue}{0.1983} & \textcolor{red}{0.1272} \\
                & FID $\downarrow$ & 215.00 & 224.35 & \textcolor{blue}{186.06} & 216.72 & 199.67 & \textcolor{red}{90.10} \\
                & Q-Align $\uparrow$ & 3.2526 & 3.5843 & 3.5282 & 3.5751 & \textcolor{blue}{3.6618} & \textcolor{red}{3.8066} \\
            \bottomrule[0.1em]
    \end{tabular}
\end{table}

\section{GPF Transferability to Other Dehazing Methods}
\label{app:gpf_transfer}

\begingroup

To evaluate its transferability, we apply GPF to
HazeFlow~\citep{shin2025hazeflow} and
Diff-Dehazer~\citep{lan2025exploiting} without retraining their
restoration networks. Table~\ref{tab:gpf_transfer_ohaze} gives the O-HAZE
results using four metrics; Tables~\ref{tab:gpf_transfer_hazeflow}--%
\ref{tab:gpf_transfer_diffdehazer} report all six metrics on all four
benchmarks, including FID and Q-Align for O-HAZE. Raw patch denotes the
original patch-based output, and + GPF denotes the final output after
global-guided fusion and detail compensation.

GPF improves PSNR, SSIM, DISTS, and FID in all eight method--dataset pairs.
LPIPS and Q-Align exhibit some method- and dataset-dependent trade-offs,
as discussed below.
Figures~\ref{fig:gpf_transfer_hazeflow_visual}
and~\ref{fig:gpf_transfer_diffdehazer_visual} complement these results with
five selected O-HAZE examples per method. These examples illustrate reduced
patch-dependent appearance variations; the tables report dataset-wide results.

\noindent \textbf{HazeFlow.}
The largest PSNR improvements occur on I-HAZE and O-HAZE, at 1.76 dB and
1.32 dB, respectively. LPIPS increases slightly on O-HAZE and Dense-Haze,
while Q-Align decreases on NH-HAZE and Dense-Haze. Thus, improved global consistency does not uniformly benefit all perceptual metrics.

\begin{table}[!t]
    \centering
    \caption{Results of applying GPF to HazeFlow on all four benchmarks.
    The best results for each metric within each dataset are highlighted in bold.}
    \label{tab:gpf_transfer_hazeflow}
    \scriptsize
    \setlength{\tabcolsep}{8.905pt}
    \begin{tabular}{ll|cccccc}
            \toprule
            \rowcolor{color-tab}
            Dataset & Setting & PSNR $\uparrow$ & SSIM $\uparrow$ & LPIPS $\downarrow$
            & DISTS $\downarrow$ & FID $\downarrow$ & Q-Align $\uparrow$ \\
            \midrule
            \multirow{2}{*}{O-HAZE} & Raw patch & 16.63 & 0.6484 & \textbf{0.4291} & 0.2054 & 152.20 & 3.3200 \\
                    & + GPF & \textbf{17.95} & \textbf{0.6782} & 0.4295 & \textbf{0.1732} & \textbf{133.54} & \textbf{3.4142} \\
            \midrule
            \multirow{2}{*}{NH-HAZE} & Raw patch & 14.38 & 0.6343 & 0.4180 & 0.2021 & 162.45 & \textbf{3.1193} \\
                    & + GPF & \textbf{14.72} & \textbf{0.6396} & \textbf{0.4105} & \textbf{0.1901} & \textbf{157.43} & 3.0556 \\
            \midrule
            \multirow{2}{*}{Dense-Haze} & Raw patch & 12.10 & 0.5320 & \textbf{0.6374} & 0.3834 & 236.11 & \textbf{2.4468} \\
                       & + GPF & \textbf{12.57} & \textbf{0.5416} & 0.6430 & \textbf{0.3828} & \textbf{225.49} & 2.4267 \\
            \midrule
            \multirow{2}{*}{I-HAZE} & Raw patch & 16.12 & 0.7675 & 0.4559 & 0.1974 & 168.12 & 3.4755 \\
                   & + GPF & \textbf{17.88} & \textbf{0.8115} & \textbf{0.4130} & \textbf{0.1453} & \textbf{131.19} & \textbf{3.7266} \\
            \bottomrule
    \end{tabular}
\end{table}

\noindent \textbf{Diff-Dehazer.}
Diff-Dehazer also benefits from GPF across all four benchmarks. LPIPS
improves on three datasets but increases slightly on O-HAZE; Q-Align
improves except on NH-HAZE. The selected visual examples highlight
reductions in block-shaped brightness differences across adjacent regions.

\begin{table}[!t]
    \centering
    \caption{Results of applying GPF to Diff-Dehazer on all four benchmarks.
    The best results for each metric within each dataset are highlighted in bold.}
    \label{tab:gpf_transfer_diffdehazer}
    \scriptsize
    \setlength{\tabcolsep}{8.905pt}
    \begin{tabular}{ll|cccccc}
            \toprule
            \rowcolor{color-tab}
            Dataset & Setting & PSNR $\uparrow$ & SSIM $\uparrow$ & LPIPS $\downarrow$
            & DISTS $\downarrow$ & FID $\downarrow$ & Q-Align $\uparrow$ \\
            \midrule
            \multirow{2}{*}{O-HAZE} & Raw patch & 16.62 & 0.6332 & \textbf{0.4085} & 0.1935 & 144.15 & 3.2273 \\
                    & + GPF & \textbf{16.78} & \textbf{0.6456} & 0.4089 & \textbf{0.1783} & \textbf{109.77} & \textbf{3.5442} \\
            \midrule
            \multirow{2}{*}{NH-HAZE} & Raw patch & 12.99 & 0.6079 & 0.3936 & 0.2094 & 151.51 & \textbf{3.4304} \\
                    & + GPF & \textbf{13.11} & \textbf{0.6156} & \textbf{0.3841} & \textbf{0.1981} & \textbf{145.86} & 3.4129 \\
            \midrule
            \multirow{2}{*}{Dense-Haze} & Raw patch & 10.94 & 0.4656 & 0.7149 & 0.5034 & 256.24 & 2.1314 \\
                       & + GPF & \textbf{11.25} & \textbf{0.4704} & \textbf{0.6946} & \textbf{0.4834} & \textbf{244.70} & \textbf{2.1626} \\
            \midrule
            \multirow{2}{*}{I-HAZE} & Raw patch & 15.83 & 0.6709 & 0.4421 & 0.1849 & 157.31 & 3.1771 \\
                   & + GPF & \textbf{16.50} & \textbf{0.6752} & \textbf{0.4293} & \textbf{0.1804} & \textbf{123.66} & \textbf{3.6387} \\
            \bottomrule
    \end{tabular}
\end{table}

\endgroup
\begingroup
\begin{figure}[!t]
    \centering
    \begingroup
    \scriptsize
    \setlength{\tabcolsep}{0.000981033\linewidth}
    \renewcommand{\arraystretch}{0}
    \begin{tabular}{@{}cccc@{}}
    \includegraphics[width=0.248528450\linewidth,height=0.165271419250\linewidth]{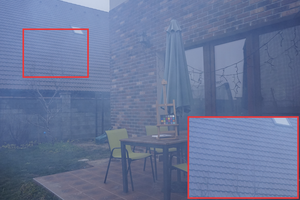} &
    \includegraphics[width=0.248528450\linewidth,height=0.165271419250\linewidth]{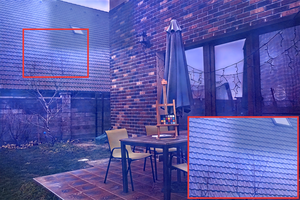} &
    \includegraphics[width=0.248528450\linewidth,height=0.165271419250\linewidth]{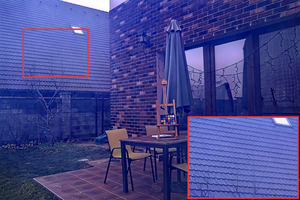} &
    \includegraphics[width=0.248528450\linewidth,height=0.165271419250\linewidth]{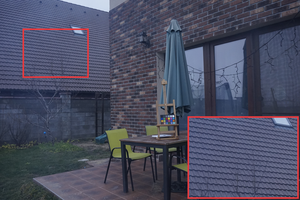} \\[1.036bp]
    \includegraphics[width=0.248528450\linewidth,height=0.175833878375\linewidth]{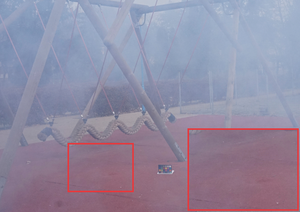} &
    \includegraphics[width=0.248528450\linewidth,height=0.175833878375\linewidth]{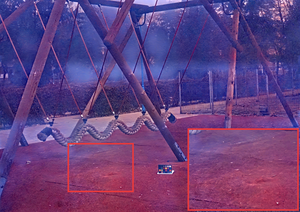} &
    \includegraphics[width=0.248528450\linewidth,height=0.175833878375\linewidth]{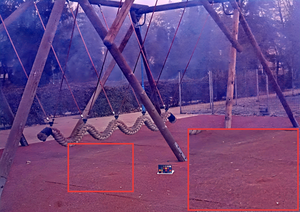} &
    \includegraphics[width=0.248528450\linewidth,height=0.175833878375\linewidth]{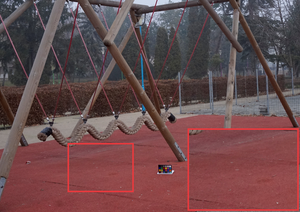} \\[1.036bp]
    \includegraphics[width=0.248528450\linewidth,height=0.213113145875\linewidth]{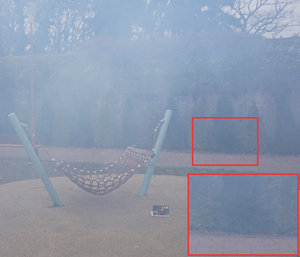} &
    \includegraphics[width=0.248528450\linewidth,height=0.213113145875\linewidth]{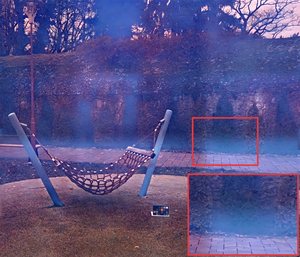} &
    \includegraphics[width=0.248528450\linewidth,height=0.213113145875\linewidth]{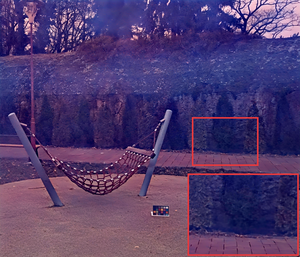} &
    \includegraphics[width=0.248528450\linewidth,height=0.213113145875\linewidth]{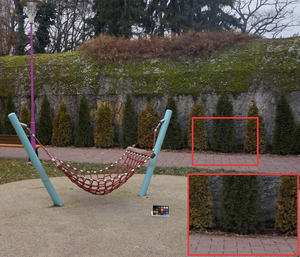} \\[1.036bp]
    \includegraphics[width=0.248528450\linewidth,height=0.165271419250\linewidth]{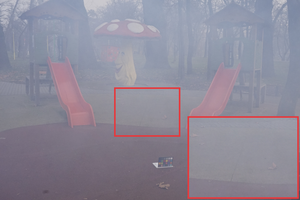} &
    \includegraphics[width=0.248528450\linewidth,height=0.165271419250\linewidth]{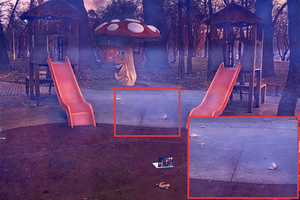} &
    \includegraphics[width=0.248528450\linewidth,height=0.165271419250\linewidth]{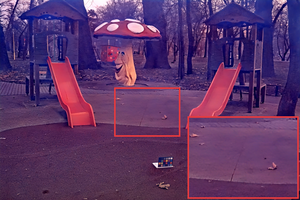} &
    \includegraphics[width=0.248528450\linewidth,height=0.165271419250\linewidth]{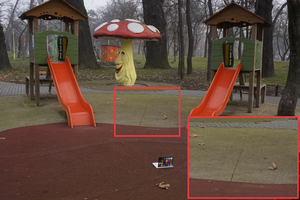} \\[1.036bp]
    \includegraphics[width=0.248528450\linewidth,height=0.150981033375\linewidth]{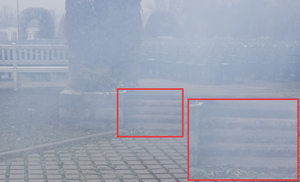} &
    \includegraphics[width=0.248528450\linewidth,height=0.150981033375\linewidth]{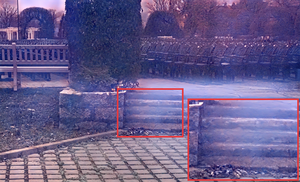} &
    \includegraphics[width=0.248528450\linewidth,height=0.150981033375\linewidth]{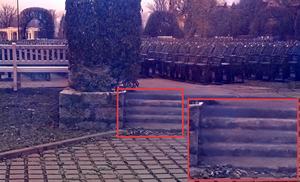} &
    \includegraphics[width=0.248528450\linewidth,height=0.150981033375\linewidth]{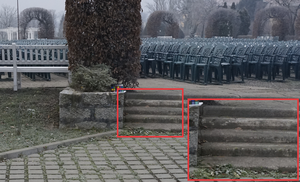} \\
    \rule{0pt}{9pt}Hazy input & HazeFlow & HazeFlow + GPF & GT \\
    \end{tabular}
    \endgroup
    \caption{Visual comparison of HazeFlow before and after applying GPF on O-HAZE.
    Rows show samples 03, 15, 17, 21, and 28, from top to bottom.
    Insets show matched crops.}
    \label{fig:gpf_transfer_hazeflow_visual}
\end{figure}

\begin{figure}[!t]
    \centering
    \begingroup
    \scriptsize
    \setlength{\tabcolsep}{0.000981033\linewidth}
    \renewcommand{\arraystretch}{0}
    \begin{tabular}{@{}cccc@{}}
    \includegraphics[width=0.248528450\linewidth,height=0.165271419250\linewidth]{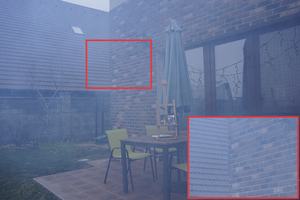} &
    \includegraphics[width=0.248528450\linewidth,height=0.165271419250\linewidth]{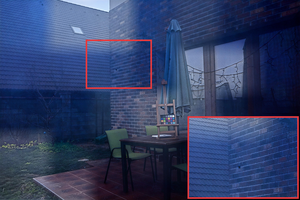} &
    \includegraphics[width=0.248528450\linewidth,height=0.165271419250\linewidth]{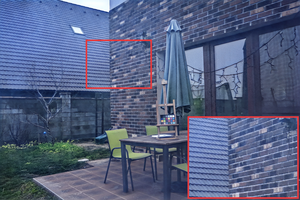} &
    \includegraphics[width=0.248528450\linewidth,height=0.165271419250\linewidth]{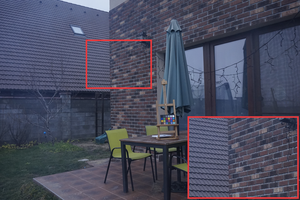} \\[1.036bp]
    \includegraphics[width=0.248528450\linewidth,height=0.162786134750\linewidth]{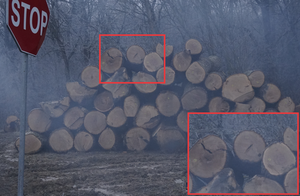} &
    \includegraphics[width=0.248528450\linewidth,height=0.162786134750\linewidth]{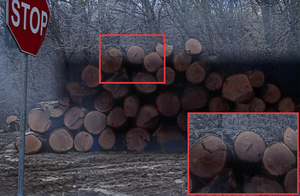} &
    \includegraphics[width=0.248528450\linewidth,height=0.162786134750\linewidth]{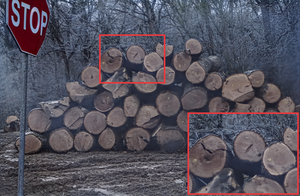} &
    \includegraphics[width=0.248528450\linewidth,height=0.162786134750\linewidth]{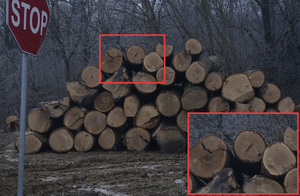} \\[1.036bp]
    \includegraphics[width=0.248528450\linewidth,height=0.165271419250\linewidth]{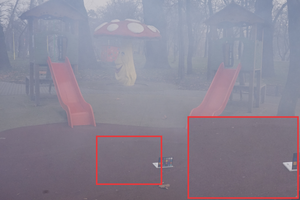} &
    \includegraphics[width=0.248528450\linewidth,height=0.165271419250\linewidth]{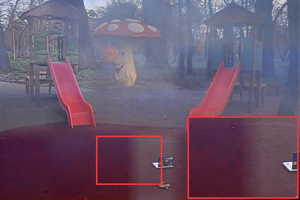} &
    \includegraphics[width=0.248528450\linewidth,height=0.165271419250\linewidth]{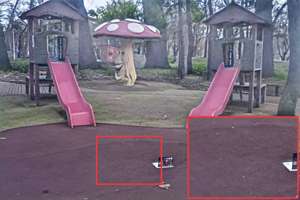} &
    \includegraphics[width=0.248528450\linewidth,height=0.165271419250\linewidth]{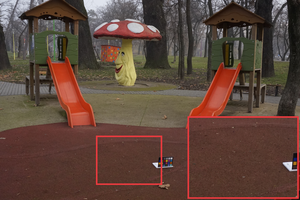} \\[1.036bp]
    \includegraphics[width=0.248528450\linewidth,height=0.188260300875\linewidth]{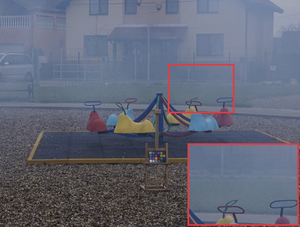} &
    \includegraphics[width=0.248528450\linewidth,height=0.188260300875\linewidth]{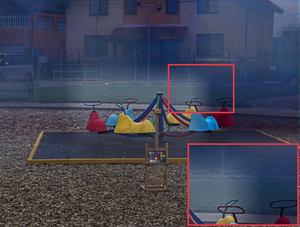} &
    \includegraphics[width=0.248528450\linewidth,height=0.188260300875\linewidth]{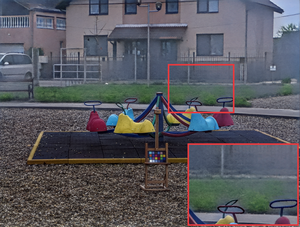} &
    \includegraphics[width=0.248528450\linewidth,height=0.188260300875\linewidth]{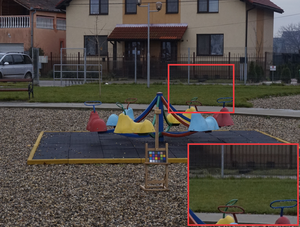} \\[1.036bp]
    \includegraphics[width=0.248528450\linewidth,height=0.188260300875\linewidth]{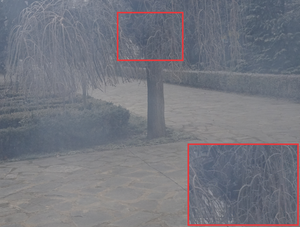} &
    \includegraphics[width=0.248528450\linewidth,height=0.188260300875\linewidth]{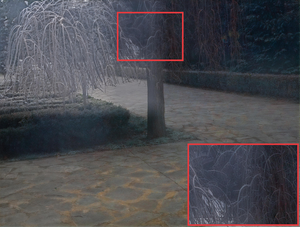} &
    \includegraphics[width=0.248528450\linewidth,height=0.188260300875\linewidth]{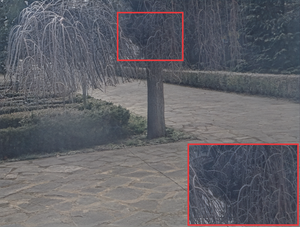} &
    \includegraphics[width=0.248528450\linewidth,height=0.188260300875\linewidth]{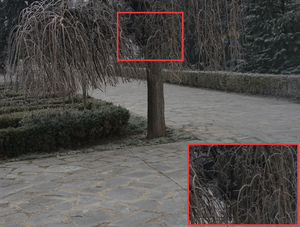} \\
    \rule{0pt}{9pt}Hazy input & Diff-Dehazer & Diff-Dehazer + GPF & GT \\
    \end{tabular}
    \endgroup
    \caption{Visual comparison of Diff-Dehazer before and after applying GPF on O-HAZE.
    Rows show samples 03, 07, 21, 24, and 33, from top to bottom.
    Insets show matched crops.}
    \label{fig:gpf_transfer_diffdehazer_visual}
\end{figure}

\endgroup

\begingroup

\newcommand{\suppFeatureImage}[2][0 0 0 0]{%
    \adjustbox{valign=t}{%
        \begin{minipage}{0.49\textwidth}
            \centering
            \includegraphics[width=\linewidth,trim=#1,clip]{figs/feature/#2}%
            \par\nointerlineskip\vspace{1pt}%
            {\scriptsize\makebox[0.5\linewidth]{Hazy input}%
            \makebox[0.5\linewidth]{SDA features}}%
        \end{minipage}%
    }%
}

\section{Additional Qualitative Results}
\label{app:additional_visual_results}

This appendix extends the comparisons in Sec.~\ref{sec:comparison} and the
SDA visualization in Sec.~\ref{sec:sda}. Full-image views across all four
benchmarks show scene-wide color, illumination, and residual haze.

\subsection{Full-Image Qualitative Comparisons}
\label{app:full_image_comparisons}

Figures~\ref{fig:supp_full_ohaze}--\ref{fig:supp_full_nhhaze} present
12 full-image comparisons, three per benchmark.
Each example contains two rows of four images: hazy input, PTTD, DiffUIR,
and Diff-Dehazer in the first row; DiffDehaze, HazeFlow, our method,
and ground truth (GT) in the second. Sample IDs appear in the GT labels.

\begin{figure}[!t]
    \centering
    \begingroup
    \scriptsize
    \setlength{\tabcolsep}{0.002136752\linewidth}
    \renewcommand{\arraystretch}{0}
    \begin{adjustbox}{max width=\linewidth,max totalheight=0.92\textheight}
    \begin{tabular}{@{}cccc@{}}
    \includegraphics[width=0.246794872\linewidth,height=0.188798077080\linewidth]{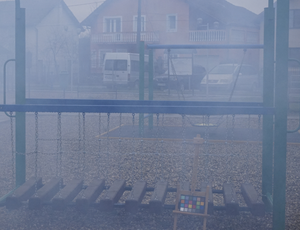} &
    \includegraphics[width=0.246794872\linewidth,height=0.188798077080\linewidth]{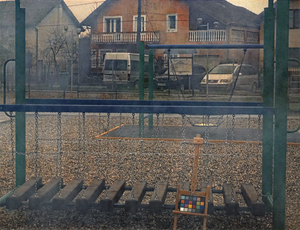} &
    \includegraphics[width=0.246794872\linewidth,height=0.188798077080\linewidth]{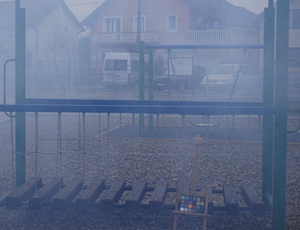} &
    \includegraphics[width=0.246794872\linewidth,height=0.188798077080\linewidth]{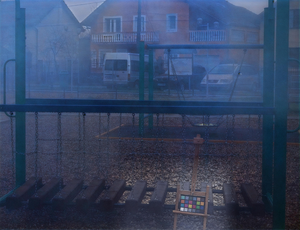} \\
    \rule{0pt}{7pt}Hazy input & PTTD & DiffUIR & Diff-Dehazer \\[4pt]
    \includegraphics[width=0.246794872\linewidth,height=0.188798077080\linewidth]{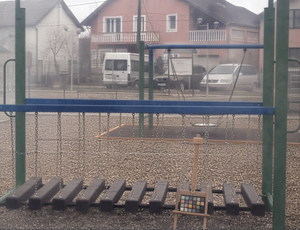} &
    \includegraphics[width=0.246794872\linewidth,height=0.188798077080\linewidth]{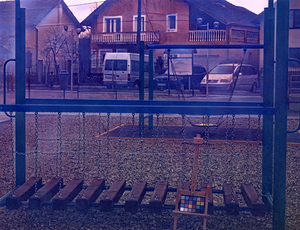} &
    \includegraphics[width=0.246794872\linewidth,height=0.188798077080\linewidth]{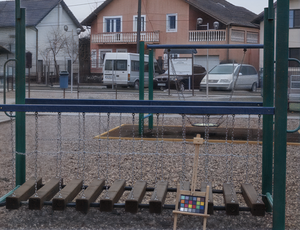} &
    \includegraphics[width=0.246794872\linewidth,height=0.188798077080\linewidth]{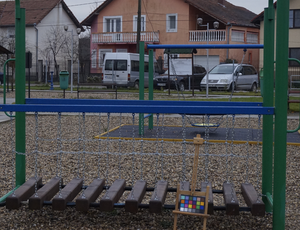} \\
    \rule{0pt}{7pt}DiffDehaze & HazeFlow & \textbf{Ours} & GT (O-HAZE 25) \\[7pt]
    \includegraphics[width=0.246794872\linewidth,height=0.212243589920\linewidth]{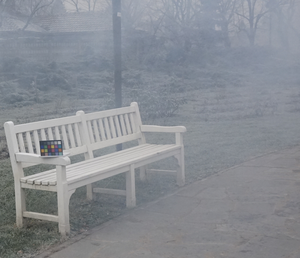} &
    \includegraphics[width=0.246794872\linewidth,height=0.212243589920\linewidth]{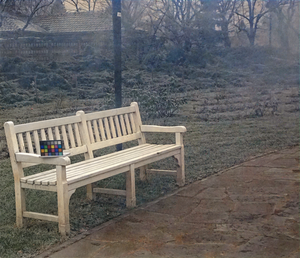} &
    \includegraphics[width=0.246794872\linewidth,height=0.212243589920\linewidth]{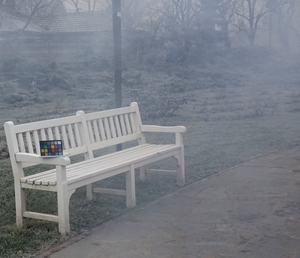} &
    \includegraphics[width=0.246794872\linewidth,height=0.212243589920\linewidth]{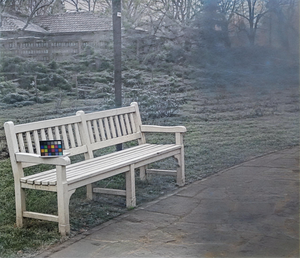} \\
    \rule{0pt}{7pt}Hazy input & PTTD & DiffUIR & Diff-Dehazer \\[4pt]
    \includegraphics[width=0.246794872\linewidth,height=0.212243589920\linewidth]{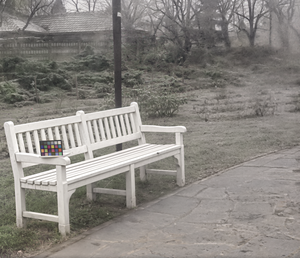} &
    \includegraphics[width=0.246794872\linewidth,height=0.212243589920\linewidth]{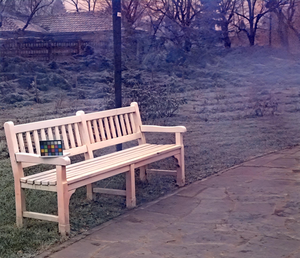} &
    \includegraphics[width=0.246794872\linewidth,height=0.212243589920\linewidth]{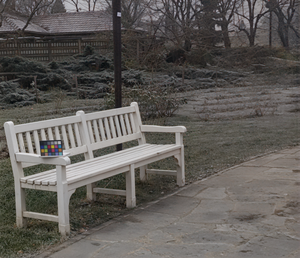} &
    \includegraphics[width=0.246794872\linewidth,height=0.212243589920\linewidth]{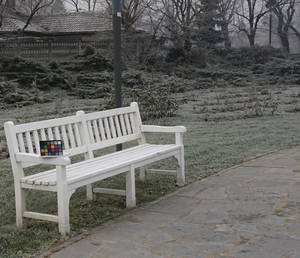} \\
    \rule{0pt}{7pt}DiffDehaze & HazeFlow & \textbf{Ours} & GT (O-HAZE 27) \\[7pt]
    \includegraphics[width=0.246794872\linewidth,height=0.179543269380\linewidth]{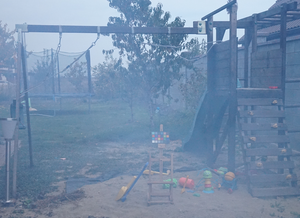} &
    \includegraphics[width=0.246794872\linewidth,height=0.179543269380\linewidth]{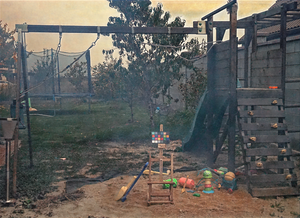} &
    \includegraphics[width=0.246794872\linewidth,height=0.179543269380\linewidth]{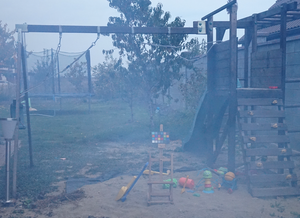} &
    \includegraphics[width=0.246794872\linewidth,height=0.179543269380\linewidth]{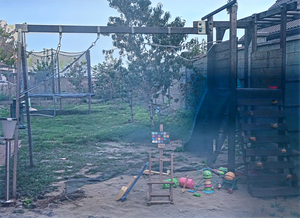} \\
    \rule{0pt}{7pt}Hazy input & PTTD & DiffUIR & Diff-Dehazer \\[4pt]
    \includegraphics[width=0.246794872\linewidth,height=0.179543269380\linewidth]{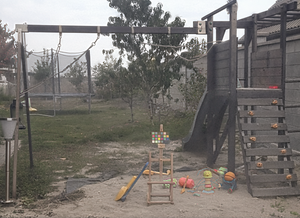} &
    \includegraphics[width=0.246794872\linewidth,height=0.179543269380\linewidth]{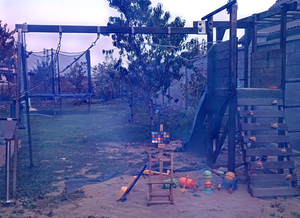} &
    \includegraphics[width=0.246794872\linewidth,height=0.179543269380\linewidth]{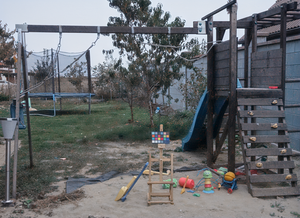} &
    \includegraphics[width=0.246794872\linewidth,height=0.179543269380\linewidth]{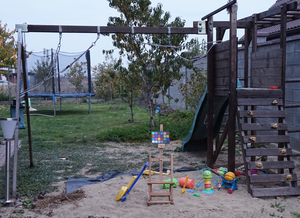} \\
    \rule{0pt}{7pt}DiffDehaze & HazeFlow & \textbf{Ours} & GT (O-HAZE 39) \\
    \end{tabular}
    \end{adjustbox}
    \endgroup
    \caption{Full-image qualitative comparisons on O-HAZE.}
    \label{fig:supp_full_ohaze}
\end{figure}

\begin{figure}[!t]
    \centering
    \begingroup
    \scriptsize
    \setlength{\tabcolsep}{0.002136752\linewidth}
    \renewcommand{\arraystretch}{0}
    \begin{adjustbox}{max width=\linewidth,max totalheight=0.92\textheight}
    \begin{tabular}{@{}cccc@{}}
    \includegraphics[width=0.246794872\linewidth,height=0.149927884740\linewidth]{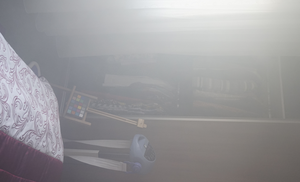} &
    \includegraphics[width=0.246794872\linewidth,height=0.149927884740\linewidth]{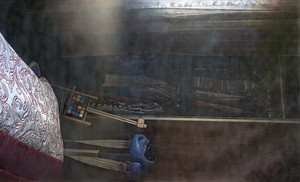} &
    \includegraphics[width=0.246794872\linewidth,height=0.149927884740\linewidth]{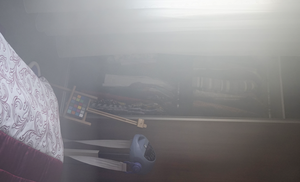} &
    \includegraphics[width=0.246794872\linewidth,height=0.149927884740\linewidth]{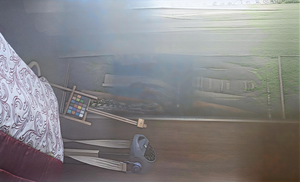} \\
    \rule{0pt}{7pt}Hazy input & PTTD & DiffUIR & Diff-Dehazer \\[4pt]
    \includegraphics[width=0.246794872\linewidth,height=0.149927884740\linewidth]{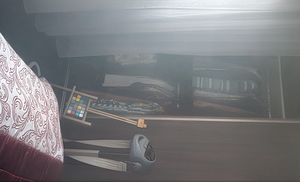} &
    \includegraphics[width=0.246794872\linewidth,height=0.149927884740\linewidth]{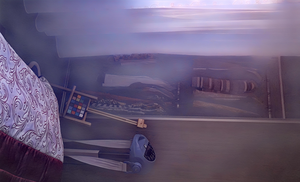} &
    \includegraphics[width=0.246794872\linewidth,height=0.149927884740\linewidth]{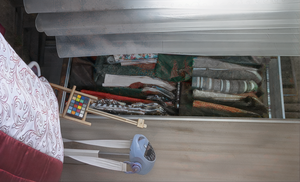} &
    \includegraphics[width=0.246794872\linewidth,height=0.149927884740\linewidth]{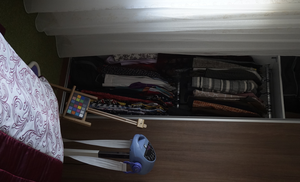} \\
    \rule{0pt}{7pt}DiffDehaze & HazeFlow & \textbf{Ours} & GT (I-HAZE 18) \\[7pt]
    \includegraphics[width=0.246794872\linewidth,height=0.168437500140\linewidth]{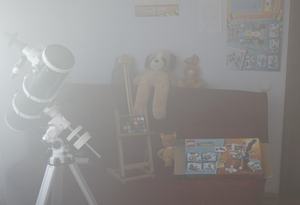} &
    \includegraphics[width=0.246794872\linewidth,height=0.168437500140\linewidth]{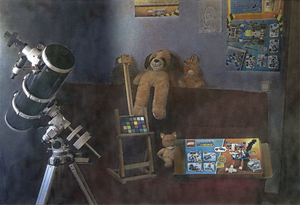} &
    \includegraphics[width=0.246794872\linewidth,height=0.168437500140\linewidth]{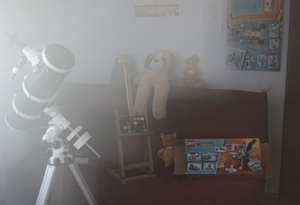} &
    \includegraphics[width=0.246794872\linewidth,height=0.168437500140\linewidth]{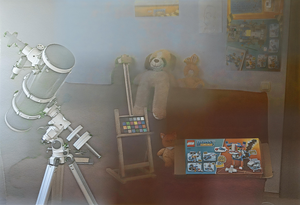} \\
    \rule{0pt}{7pt}Hazy input & PTTD & DiffUIR & Diff-Dehazer \\[4pt]
    \includegraphics[width=0.246794872\linewidth,height=0.168437500140\linewidth]{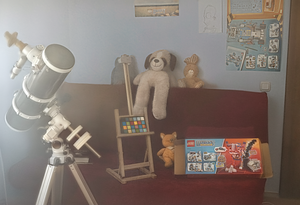} &
    \includegraphics[width=0.246794872\linewidth,height=0.168437500140\linewidth]{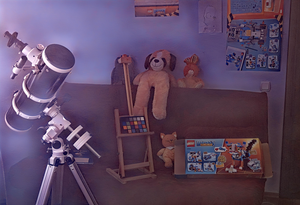} &
    \includegraphics[width=0.246794872\linewidth,height=0.168437500140\linewidth]{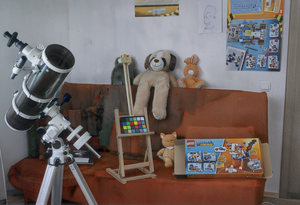} &
    \includegraphics[width=0.246794872\linewidth,height=0.168437500140\linewidth]{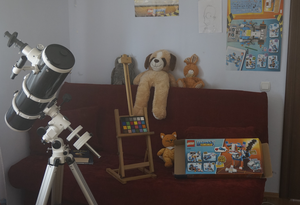} \\
    \rule{0pt}{7pt}DiffDehaze & HazeFlow & \textbf{Ours} & GT (I-HAZE 34) \\[7pt]
    \includegraphics[width=0.246794872\linewidth,height=0.161650641160\linewidth]{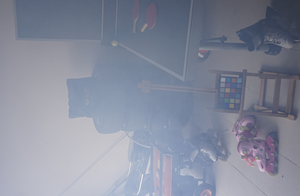} &
    \includegraphics[width=0.246794872\linewidth,height=0.161650641160\linewidth]{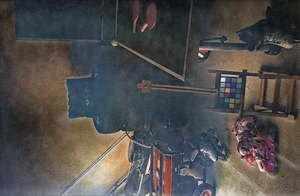} &
    \includegraphics[width=0.246794872\linewidth,height=0.161650641160\linewidth]{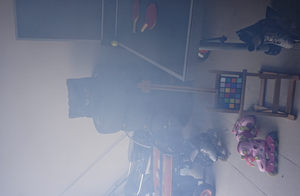} &
    \includegraphics[width=0.246794872\linewidth,height=0.161650641160\linewidth]{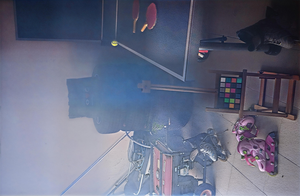} \\
    \rule{0pt}{7pt}Hazy input & PTTD & DiffUIR & Diff-Dehazer \\[4pt]
    \includegraphics[width=0.246794872\linewidth,height=0.161650641160\linewidth]{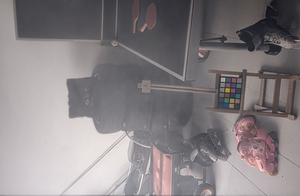} &
    \includegraphics[width=0.246794872\linewidth,height=0.161650641160\linewidth]{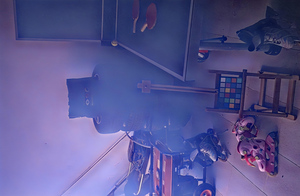} &
    \includegraphics[width=0.246794872\linewidth,height=0.161650641160\linewidth]{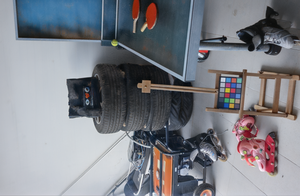} &
    \includegraphics[width=0.246794872\linewidth,height=0.161650641160\linewidth]{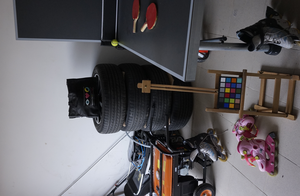} \\
    \rule{0pt}{7pt}DiffDehaze & HazeFlow & \textbf{Ours} & GT (I-HAZE 35) \\
    \end{tabular}
    \end{adjustbox}
    \endgroup
    \caption{Full-image qualitative comparisons on I-HAZE.}
    \label{fig:supp_full_ihaze}
\end{figure}

\begin{figure}[!t]
    \centering
    \begingroup
    \scriptsize
    \setlength{\tabcolsep}{0.002136752\linewidth}
    \renewcommand{\arraystretch}{0}
    \begin{adjustbox}{max width=\linewidth,max totalheight=0.92\textheight}
    \begin{tabular}{@{}cccc@{}}
    \includegraphics[width=0.246794872\linewidth,height=0.185096154000\linewidth]{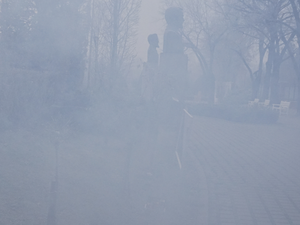} &
    \includegraphics[width=0.246794872\linewidth,height=0.185096154000\linewidth]{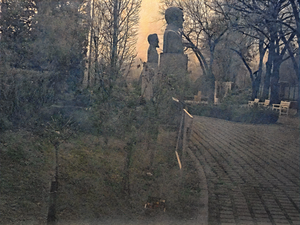} &
    \includegraphics[width=0.246794872\linewidth,height=0.185096154000\linewidth]{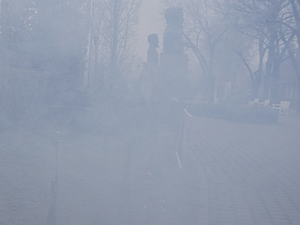} &
    \includegraphics[width=0.246794872\linewidth,height=0.185096154000\linewidth]{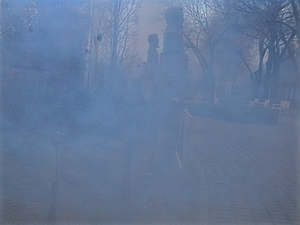} \\
    \rule{0pt}{7pt}Hazy input & PTTD & DiffUIR & Diff-Dehazer \\[4pt]
    \includegraphics[width=0.246794872\linewidth,height=0.185096154000\linewidth]{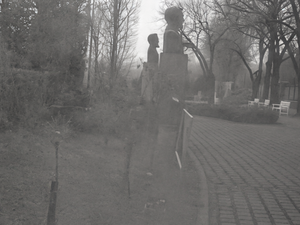} &
    \includegraphics[width=0.246794872\linewidth,height=0.185096154000\linewidth]{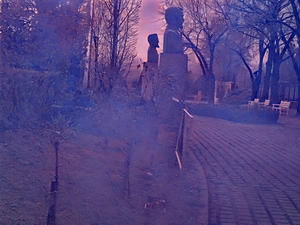} &
    \includegraphics[width=0.246794872\linewidth,height=0.185096154000\linewidth]{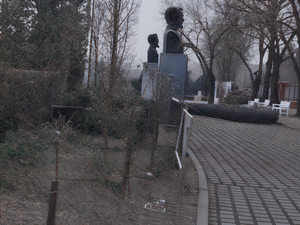} &
    \includegraphics[width=0.246794872\linewidth,height=0.185096154000\linewidth]{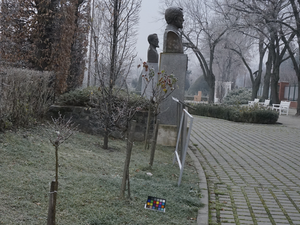} \\
    \rule{0pt}{7pt}DiffDehaze & HazeFlow & \textbf{Ours} & GT (Dense-Haze 07) \\[7pt]
    \includegraphics[width=0.246794872\linewidth,height=0.185096154000\linewidth]{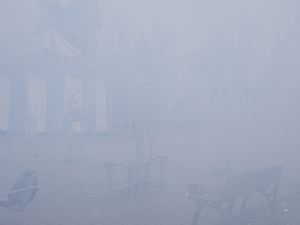} &
    \includegraphics[width=0.246794872\linewidth,height=0.185096154000\linewidth]{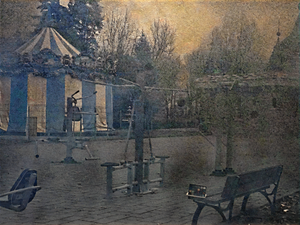} &
    \includegraphics[width=0.246794872\linewidth,height=0.185096154000\linewidth]{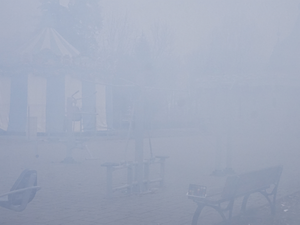} &
    \includegraphics[width=0.246794872\linewidth,height=0.185096154000\linewidth]{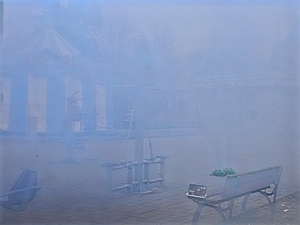} \\
    \rule{0pt}{7pt}Hazy input & PTTD & DiffUIR & Diff-Dehazer \\[4pt]
    \includegraphics[width=0.246794872\linewidth,height=0.185096154000\linewidth]{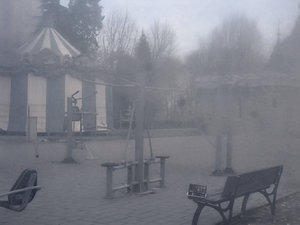} &
    \includegraphics[width=0.246794872\linewidth,height=0.185096154000\linewidth]{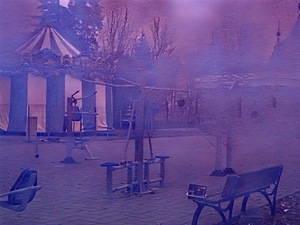} &
    \includegraphics[width=0.246794872\linewidth,height=0.185096154000\linewidth]{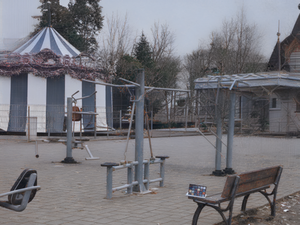} &
    \includegraphics[width=0.246794872\linewidth,height=0.185096154000\linewidth]{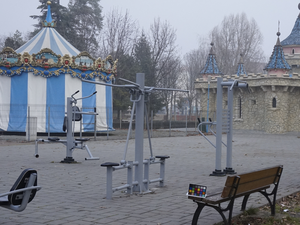} \\
    \rule{0pt}{7pt}DiffDehaze & HazeFlow & \textbf{Ours} & GT (Dense-Haze 12) \\[7pt]
    \includegraphics[width=0.246794872\linewidth,height=0.185096154000\linewidth]{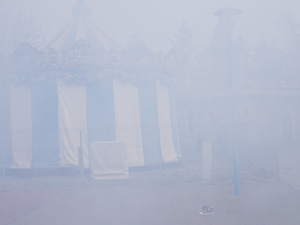} &
    \includegraphics[width=0.246794872\linewidth,height=0.185096154000\linewidth]{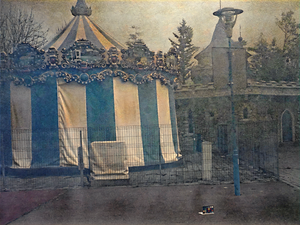} &
    \includegraphics[width=0.246794872\linewidth,height=0.185096154000\linewidth]{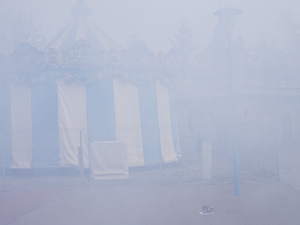} &
    \includegraphics[width=0.246794872\linewidth,height=0.185096154000\linewidth]{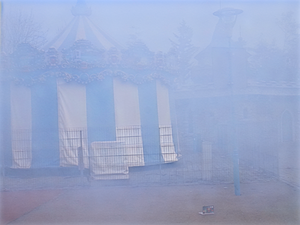} \\
    \rule{0pt}{7pt}Hazy input & PTTD & DiffUIR & Diff-Dehazer \\[4pt]
    \includegraphics[width=0.246794872\linewidth,height=0.185096154000\linewidth]{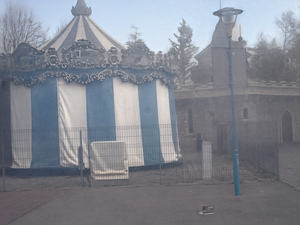} &
    \includegraphics[width=0.246794872\linewidth,height=0.185096154000\linewidth]{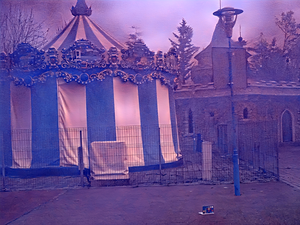} &
    \includegraphics[width=0.246794872\linewidth,height=0.185096154000\linewidth]{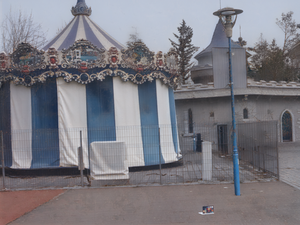} &
    \includegraphics[width=0.246794872\linewidth,height=0.185096154000\linewidth]{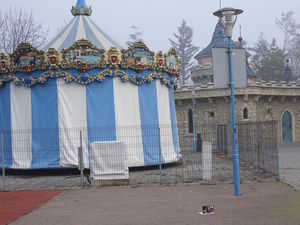} \\
    \rule{0pt}{7pt}DiffDehaze & HazeFlow & \textbf{Ours} & GT (Dense-Haze 47) \\
    \end{tabular}
    \end{adjustbox}
    \endgroup
    \caption{Full-image qualitative comparisons on Dense-Haze.}
    \label{fig:supp_full_densehaze}
\end{figure}

\begin{figure}[!t]
    \centering
    \begingroup
    \scriptsize
    \setlength{\tabcolsep}{0.002136752\linewidth}
    \renewcommand{\arraystretch}{0}
    \begin{adjustbox}{max width=\linewidth,max totalheight=0.92\textheight}
    \begin{tabular}{@{}cccc@{}}
    \includegraphics[width=0.246794872\linewidth,height=0.185096154000\linewidth]{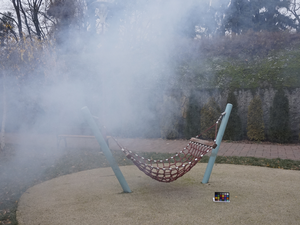} &
    \includegraphics[width=0.246794872\linewidth,height=0.185096154000\linewidth]{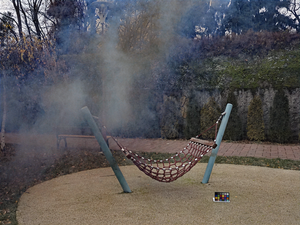} &
    \includegraphics[width=0.246794872\linewidth,height=0.185096154000\linewidth]{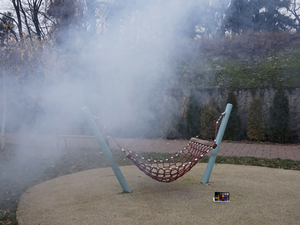} &
    \includegraphics[width=0.246794872\linewidth,height=0.185096154000\linewidth]{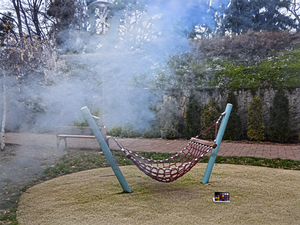} \\
    \rule{0pt}{7pt}Hazy input & PTTD & DiffUIR & Diff-Dehazer \\[4pt]
    \includegraphics[width=0.246794872\linewidth,height=0.185096154000\linewidth]{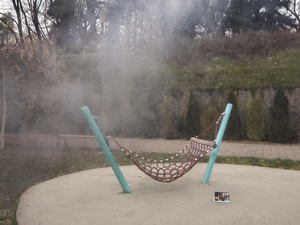} &
    \includegraphics[width=0.246794872\linewidth,height=0.185096154000\linewidth]{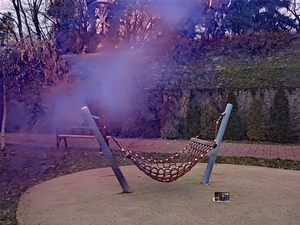} &
    \includegraphics[width=0.246794872\linewidth,height=0.185096154000\linewidth]{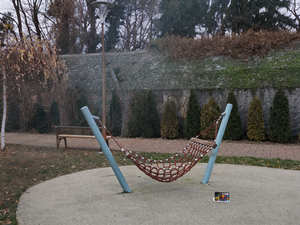} &
    \includegraphics[width=0.246794872\linewidth,height=0.185096154000\linewidth]{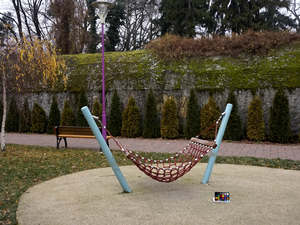} \\
    \rule{0pt}{7pt}DiffDehaze & HazeFlow & \textbf{Ours} & GT (NH-HAZE 21) \\[7pt]
    \includegraphics[width=0.246794872\linewidth,height=0.185096154000\linewidth]{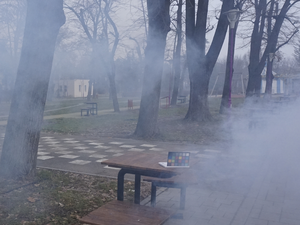} &
    \includegraphics[width=0.246794872\linewidth,height=0.185096154000\linewidth]{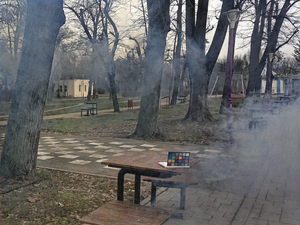} &
    \includegraphics[width=0.246794872\linewidth,height=0.185096154000\linewidth]{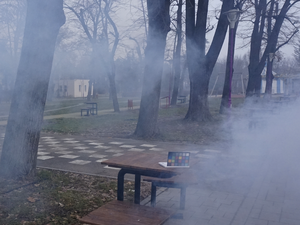} &
    \includegraphics[width=0.246794872\linewidth,height=0.185096154000\linewidth]{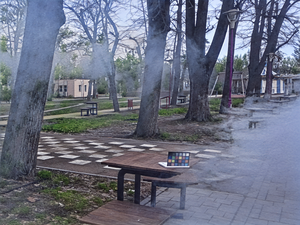} \\
    \rule{0pt}{7pt}Hazy input & PTTD & DiffUIR & Diff-Dehazer \\[4pt]
    \includegraphics[width=0.246794872\linewidth,height=0.185096154000\linewidth]{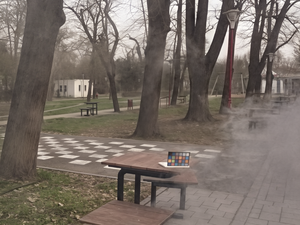} &
    \includegraphics[width=0.246794872\linewidth,height=0.185096154000\linewidth]{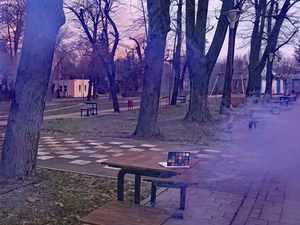} &
    \includegraphics[width=0.246794872\linewidth,height=0.185096154000\linewidth]{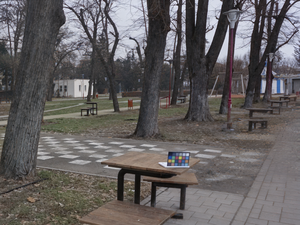} &
    \includegraphics[width=0.246794872\linewidth,height=0.185096154000\linewidth]{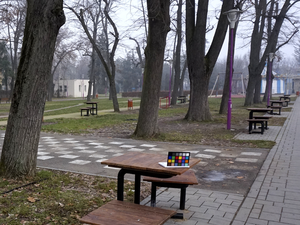} \\
    \rule{0pt}{7pt}DiffDehaze & HazeFlow & \textbf{Ours} & GT (NH-HAZE 18) \\[7pt]
    \includegraphics[width=0.246794872\linewidth,height=0.185096154000\linewidth]{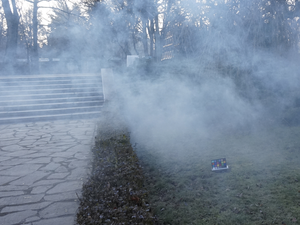} &
    \includegraphics[width=0.246794872\linewidth,height=0.185096154000\linewidth]{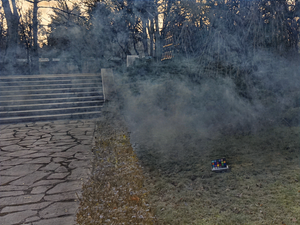} &
    \includegraphics[width=0.246794872\linewidth,height=0.185096154000\linewidth]{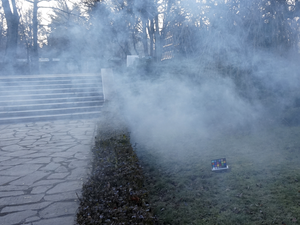} &
    \includegraphics[width=0.246794872\linewidth,height=0.185096154000\linewidth]{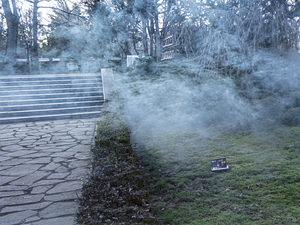} \\
    \rule{0pt}{7pt}Hazy input & PTTD & DiffUIR & Diff-Dehazer \\[4pt]
    \includegraphics[width=0.246794872\linewidth,height=0.185096154000\linewidth]{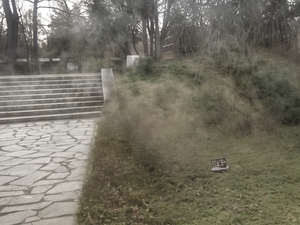} &
    \includegraphics[width=0.246794872\linewidth,height=0.185096154000\linewidth]{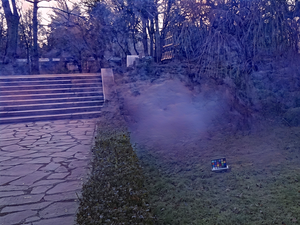} &
    \includegraphics[width=0.246794872\linewidth,height=0.185096154000\linewidth]{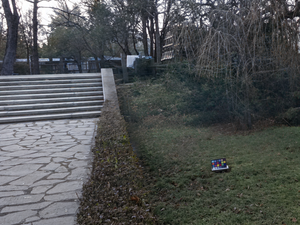} &
    \includegraphics[width=0.246794872\linewidth,height=0.185096154000\linewidth]{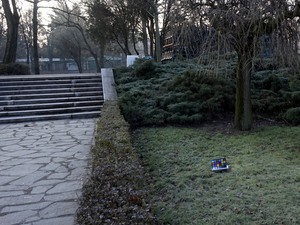} \\
    \rule{0pt}{7pt}DiffDehaze & HazeFlow & \textbf{Ours} & GT (NH-HAZE 42) \\
    \end{tabular}
    \end{adjustbox}
    \endgroup
    \caption{Full-image qualitative comparisons on NH-HAZE.}
    \label{fig:supp_full_nhhaze}
\end{figure}

\subsection{SDA Feature Visualizations}
\label{app:sda_feature_visualizations}

Figure~\ref{fig:supp_sda_features} extends the visualization in
Sec.~\ref{sec:sda} with eight NH-HAZE examples and two O-HAZE examples.
Each pair shows the hazy RGB input on the left and the corresponding SDA
feature visualization on the right, allowing the spatial structure of the
learned representation to be inspected alongside the observed haze.
The examples cover different scene structures and haze distributions,
providing a broader view of the representation than the two examples
in the main paper. Feature responses depend on both image content and
degradation, so the maps are interpreted as learned restoration cues
rather than calibrated measurements of physical haze density.

\begin{figure}[!t]
    \centering
    \setlength{\tabcolsep}{0pt}
    \renewcommand{\arraystretch}{0.85}
    \begin{tabular}{@{}c@{\hspace{0.02\textwidth}}c@{}}
        \suppFeatureImage{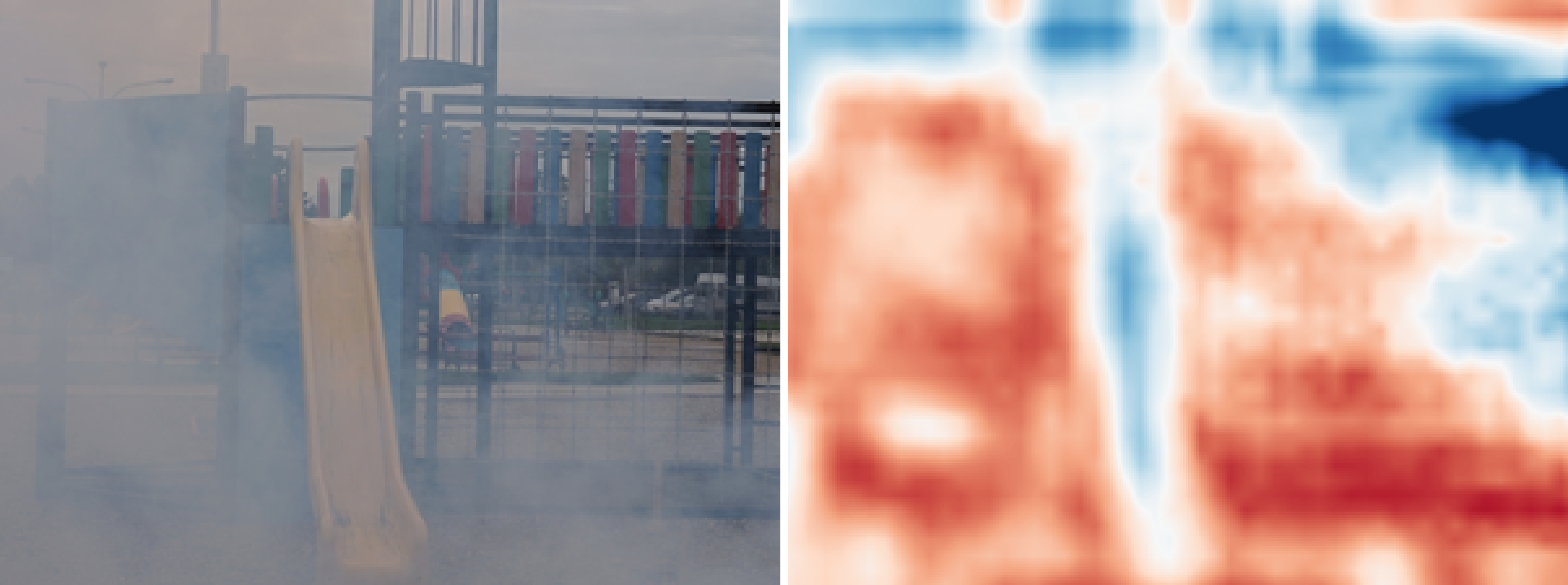} &
        \suppFeatureImage{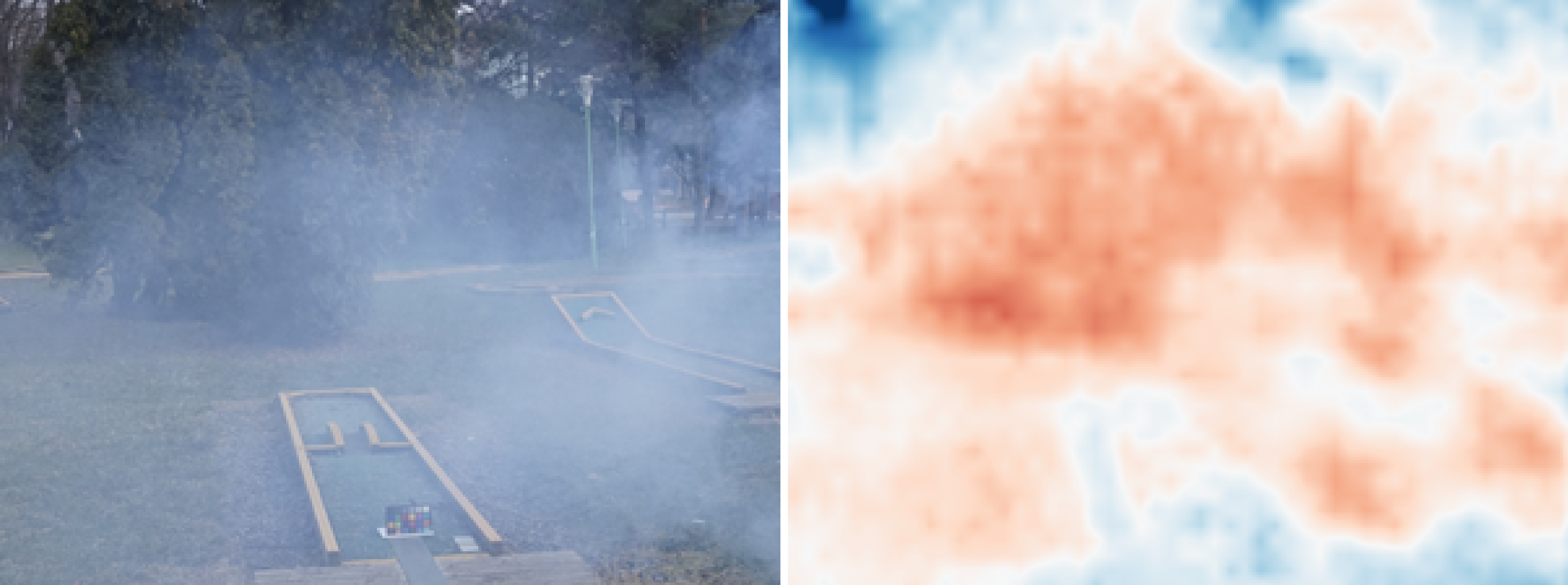} \\
        {\scriptsize NH-HAZE: 09} & {\scriptsize NH-HAZE: 12} \\[3pt]
        \suppFeatureImage{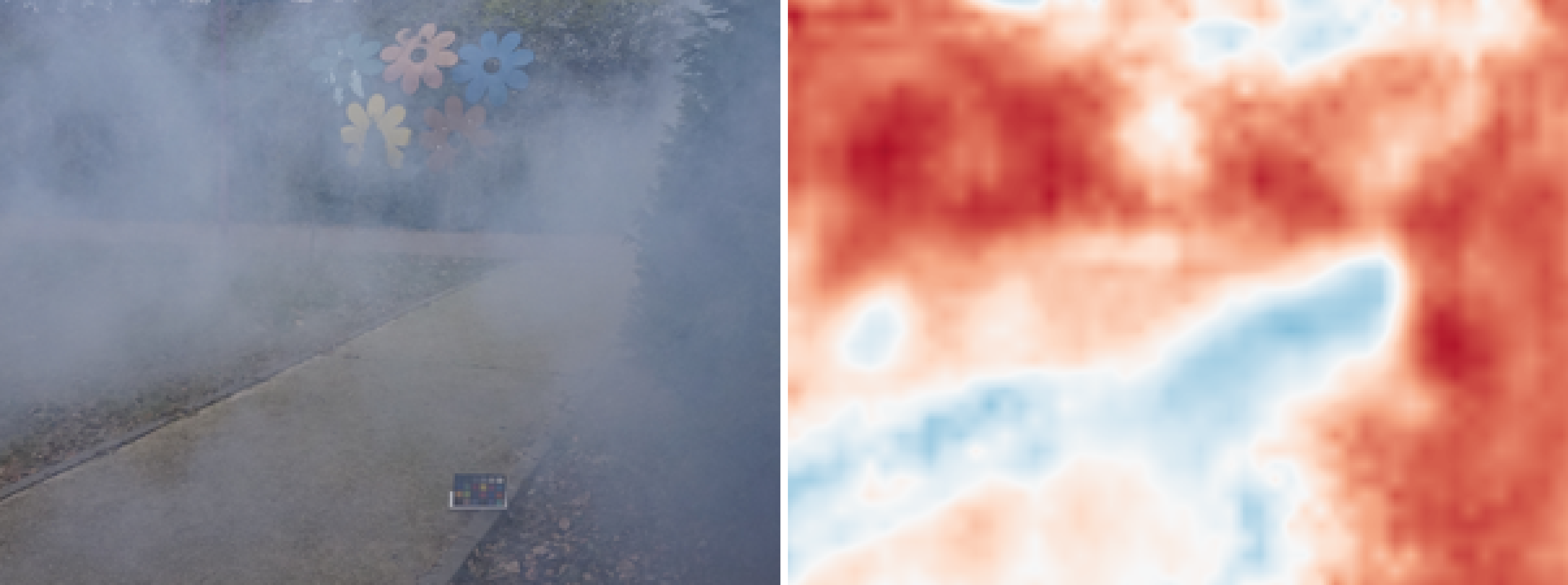} &
        \suppFeatureImage{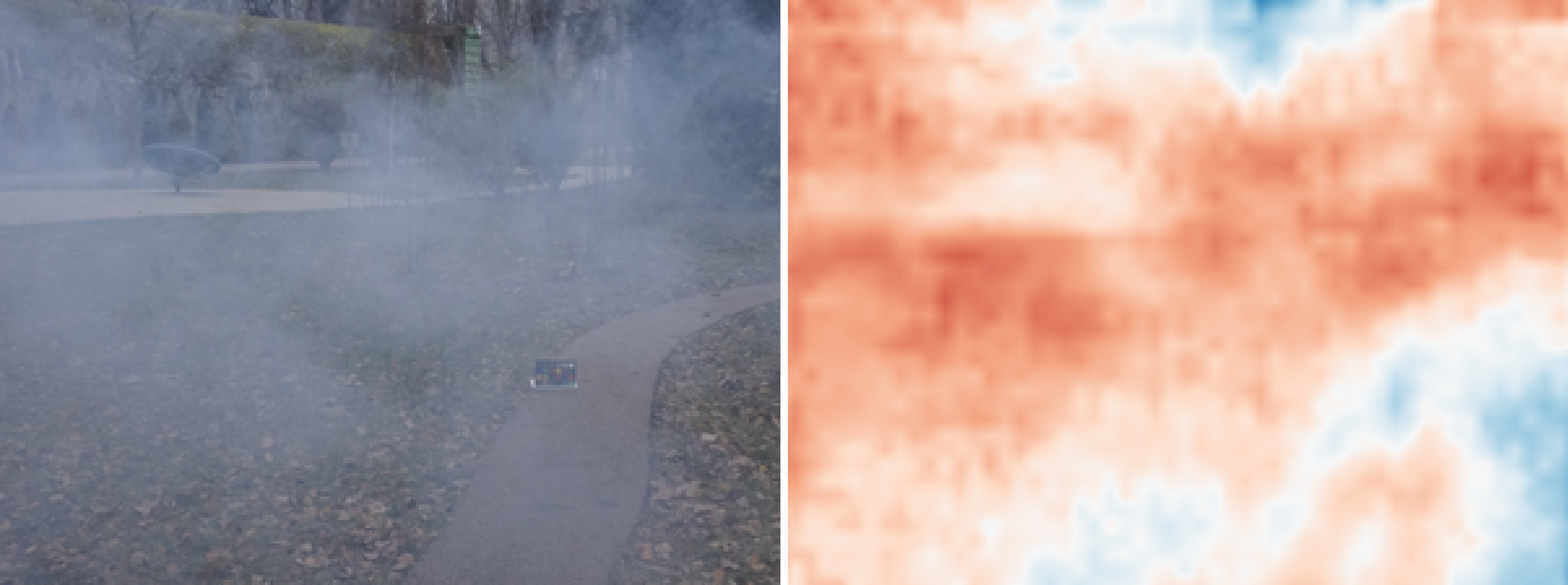} \\
        {\scriptsize NH-HAZE: 20} & {\scriptsize NH-HAZE: 27} \\[3pt]
        \suppFeatureImage{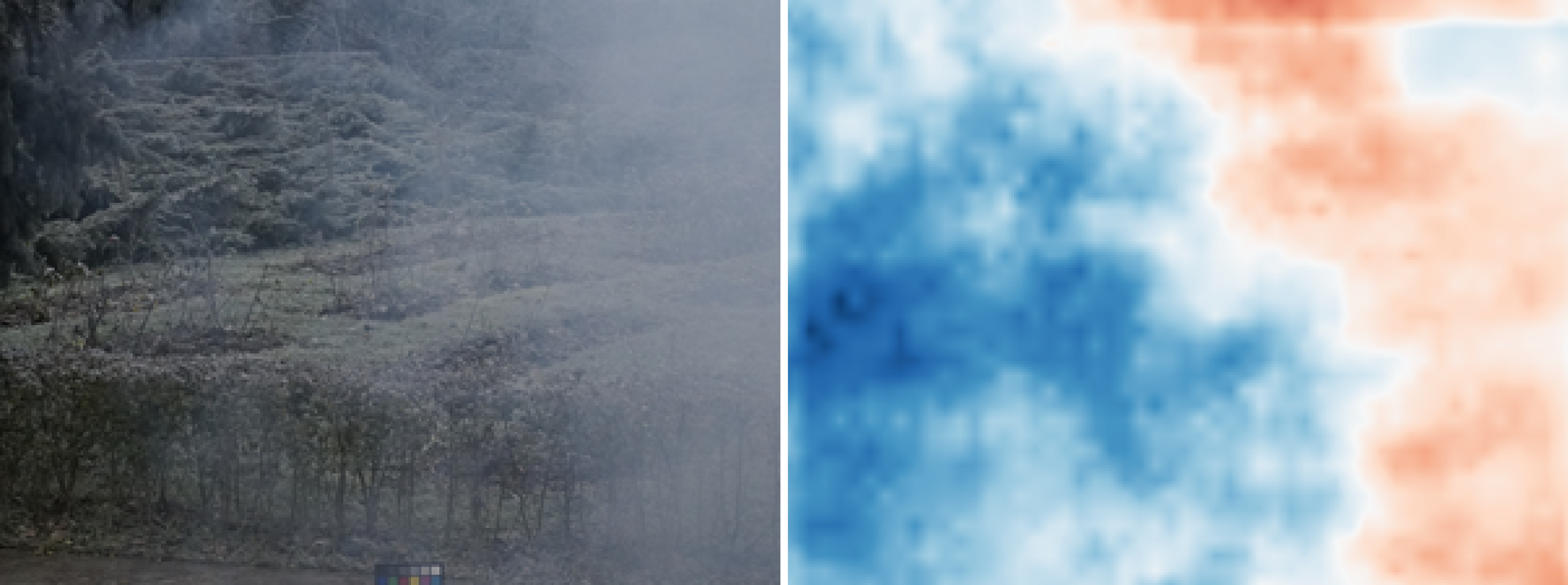} &
        \suppFeatureImage{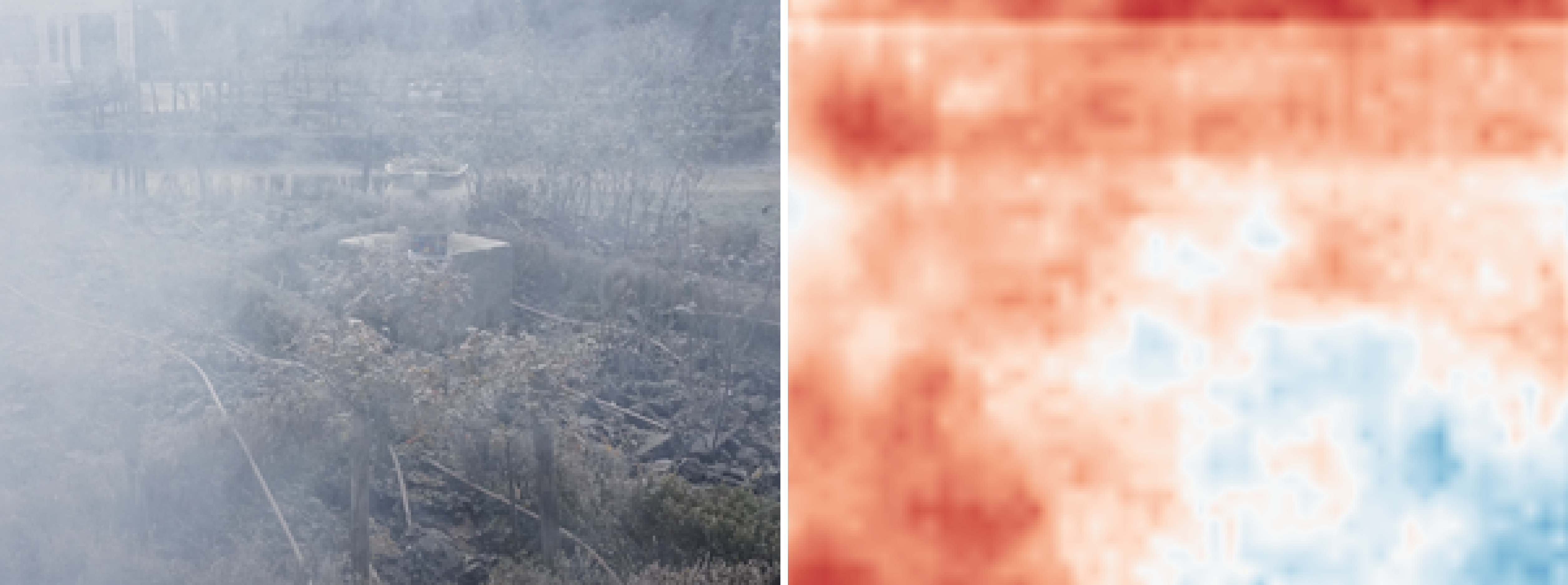} \\
        {\scriptsize NH-HAZE: 31} & {\scriptsize NH-HAZE: 34} \\[3pt]
        \suppFeatureImage{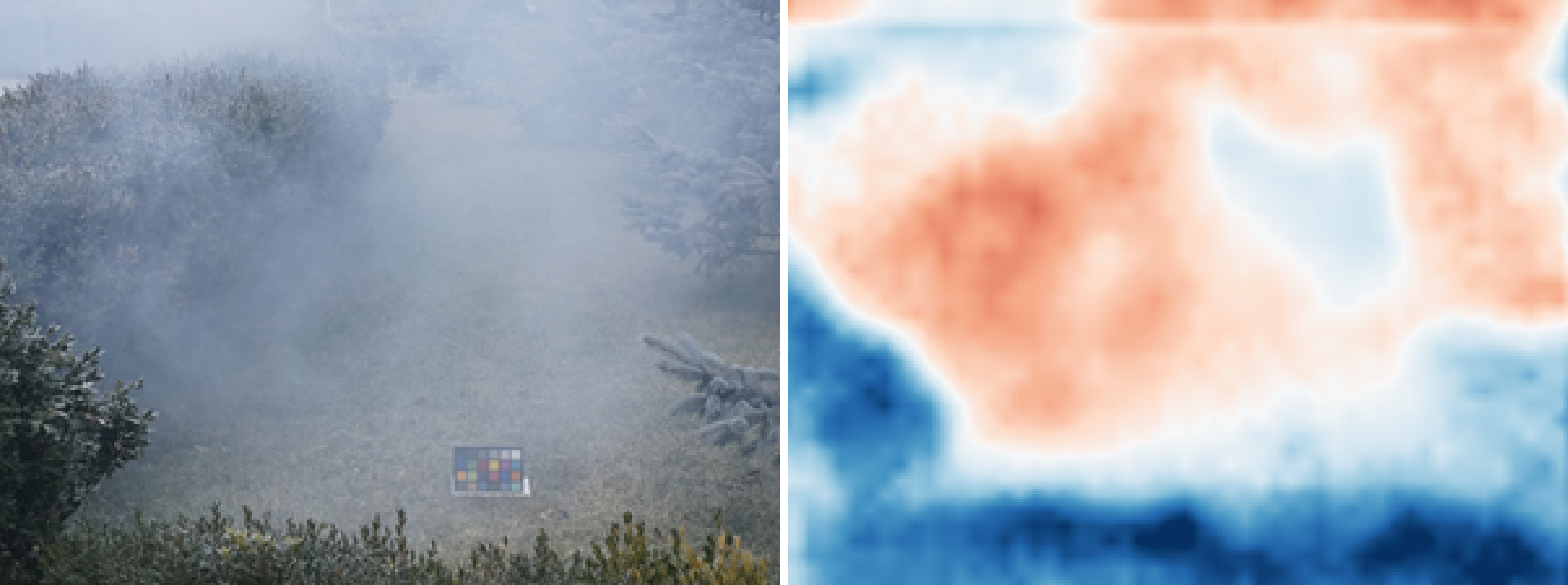} &
        \suppFeatureImage{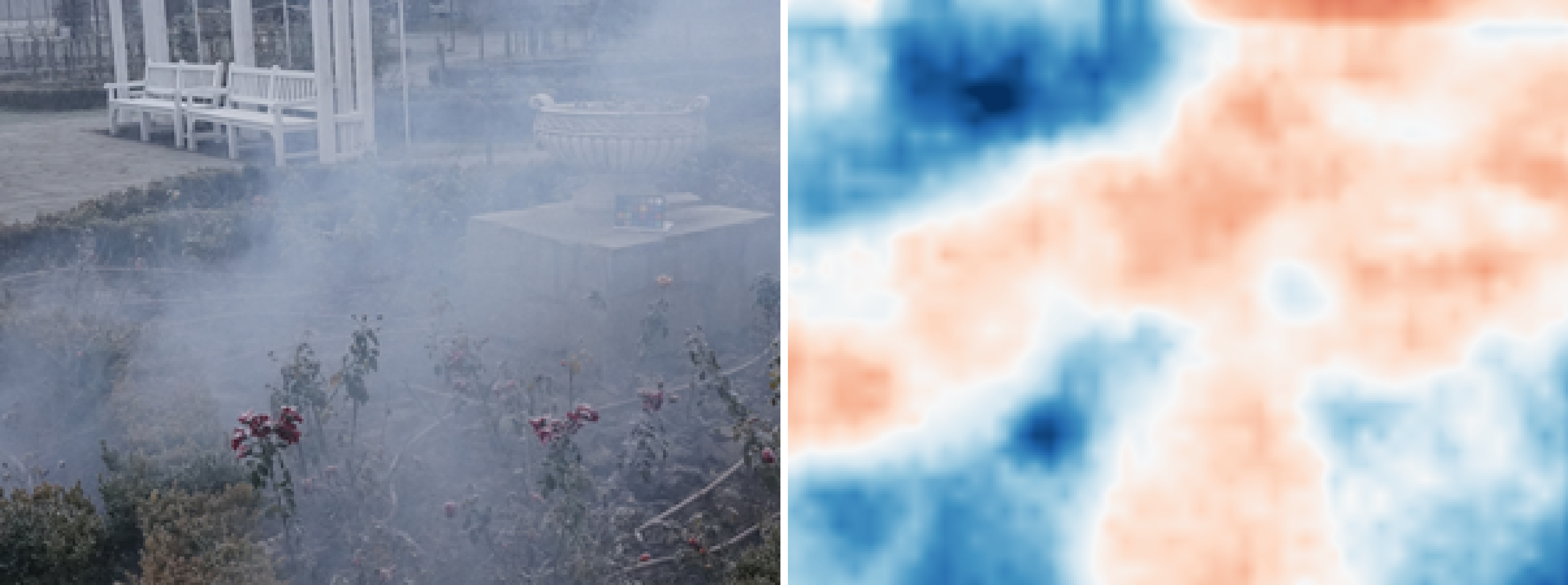} \\
        {\scriptsize NH-HAZE: 54} & {\scriptsize NH-HAZE: 55} \\[3pt]
        \suppFeatureImage{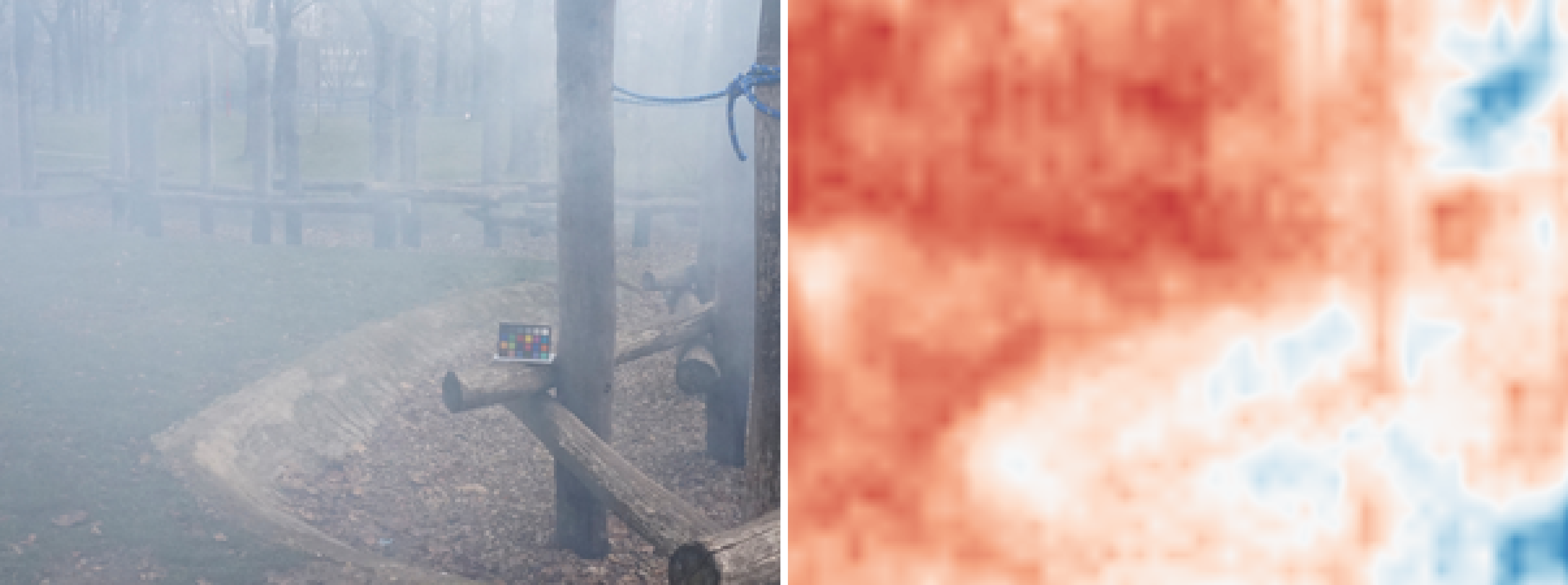} &
        \suppFeatureImage{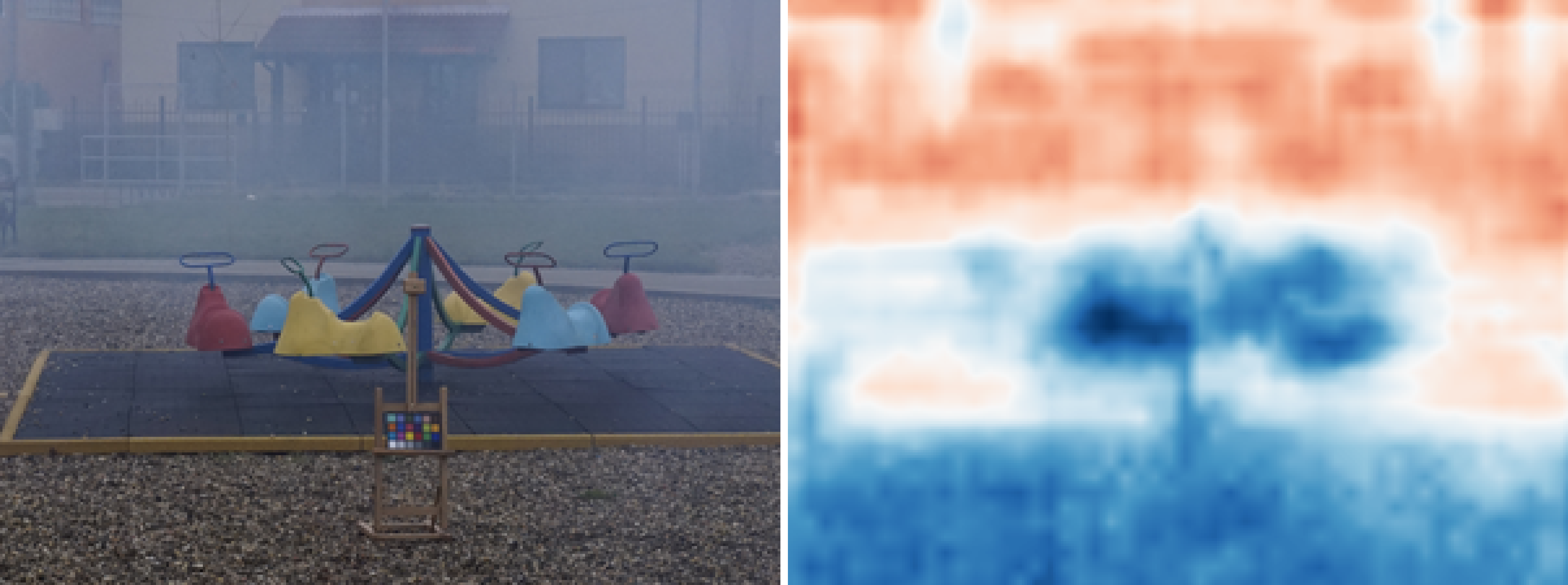} \\
        {\scriptsize O-HAZE: 20} & {\scriptsize O-HAZE: 24}
    \end{tabular}
    \caption{SDA feature visualizations on NH-HAZE and O-HAZE.
    Each pair shows the hazy input (left) and the corresponding learned feature map (right).}
    \label{fig:supp_sda_features}
\end{figure}

\endgroup

\clearpage
\endgroup
\end{document}